%% file: main.tex
\documentclass[10pt,twocolumn,letterpaper]{article}

\usepackage[pagenumbers]{iccv} 
\input{def}
\usepackage{makecell}
\usepackage{float}
\usepackage{placeins}
\usepackage{xcolor}

\definecolor{iccvblue}{rgb}{0.21,0.49,0.74}
\usepackage[pagebackref,breaklinks,colorlinks,allcolors=iccvblue]{hyperref}

\title{VoiceCraft-Dub: Automated Video Dubbing with Neural Codec Language Models}

\author{Kim Sung-Bin${}^{1}$ \,
    Jeongsoo Choi${}^{2}$ \,
    Puyuan Peng${}^{3}$ \,
    Joon Son Chung${}^{2}$ \,
    Tae-Hyun Oh${}^{4}$ \,
    David Harwath${}^{3}$ \,\\
    ${}^{1}$Department of Electrical Engineering, POSTECH\\
    ${}^{2}$School of Electrical Engineering and ${}^{4}$School of Computing, KAIST\\
    ${}^{3}$Department of Computer Science, The University of Texas at Austin\\
}

\begin{document}
\twocolumn[{
\renewcommand\twocolumn[1][]{#1}%
\maketitle
\centering
\vspace{-5.3mm}
    \centering
    \captionsetup{type=figure}
    \includegraphics[width=1\textwidth]{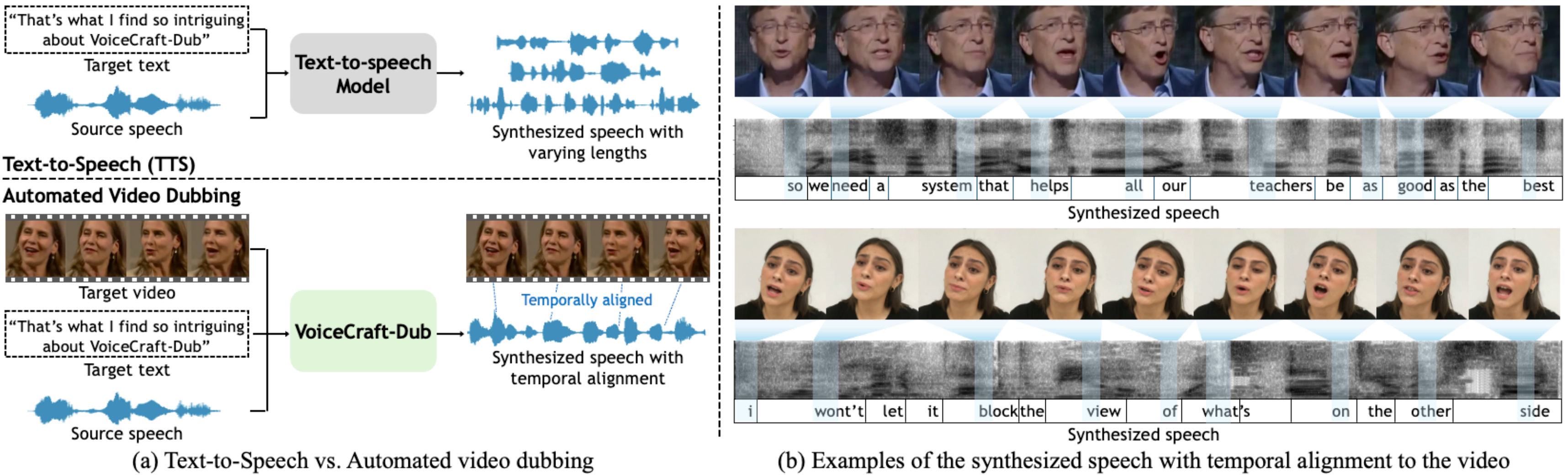}
    \vspace{-6.2mm}
    \captionof{figure}{{\bf Automated video dubbing.} (a) Unlike text-to-speech, which generates diverse speech based on target text, automated video dubbing requires synthesized speech to be temporally and expressively aligned with the video while maintaining naturalness and intelligibility. (b) Examples of synthesized speech from VoiceCraft-Dub show that each speech is aligned with the lip movements of the input video. \uline{\textbf{We strongly encourage listening to each of the synthesized samples in~\url{https://voicecraft-dub.github.io/}.}}}
    \vspace{4.2mm}
    \label{fig:teaser}
}]

\input{sec/0_abstract}    
\input{sec/1_intro}
\input{sec/2_related}

\input{sec/3_method}
\input{sec/4_exp}
\input{sec/5_conclusion}

{
    \small
    \bibliographystyle{ieeenat_fullname}
    \bibliography{main}
}

\clearpage
\appendix
\section*{Appendix}
\input{supplement}

\end{document}

%% file: def.tex
\usepackage{enumitem}
\usepackage{multirow}
\usepackage{wrapfig}
\usepackage{colortbl}
\usepackage[normalem]{ulem}
\usepackage{microtype}
\usepackage{comment}
\usepackage[hang,flushmargin]{footmisc}
\usepackage{graphicx}
\usepackage{multirow}
\setlist[itemize]{align=parleft,left=0pt,topsep=1mm,itemsep=0mm,parsep=1mm}
\definecolor{azure}{rgb}{0.0, 0.5, 1.0}
\definecolor{azure(colorwheel)}{rgb}{0.0, 0.5, 1.0}
\definecolor{nicegreen}{rgb}{0.0, 0.7, 0.1}
\definecolor{yw}{rgb}{0.01176, 0.5490, 0.5490}
\definecolor{ashblue}{rgb}{0.36, 0.54, 0.66}
\definecolor{ashgrey}{rgb}{0.7, 0.75, 0.71}
\definecolor{applegreen}{rgb}{0.55, 0.71, 0.0}
\definecolor{blue}{rgb}{0.0, 0.0, 1.0}
\definecolor{postechred}{rgb}{0.784, 0.003, 0.313}
\definecolor{ywg}{rgb}{0.9960, 0.8984, 0.5859}
\definecolor{ballblue}{rgb}{0.13, 0.67, 0.8}
\definecolor{cornellred}{rgb}{0.7, 0.11, 0.11}
\definecolor{darkcyan}{rgb}{0.0, 0.55, 0.55}
\definecolor{CuGray}{gray}{0.9}
\definecolor{airforceblue}{rgb}{0.36, 0.54, 0.66}
\definecolor{rev}{rgb}{0.784, 0.003, 0.313}
\definecolor{pink}{cmyk}{0, 0.7808, 0.4429, 0.1412}
\definecolor{amethyst}{rgb}{0.6, 0.4, 0.8}
\definecolor{black}{rgb}{0.0, 0.0, 0.0}
\definecolor{tb3_yellow}{rgb}{0.996, 1.0, 0.6}
\definecolor{tb3_orange}{rgb}{0.980, 0.8, 0.604}
\definecolor{tb3_red}{rgb}{0.972, 0.6, 0.6}
\definecolor{dimgray}{rgb}{0.41, 0.41, 0.41}
\definecolor{brickred}{rgb}{0.8, 0.25, 0.33}
\definecolor{bleudefrance}{rgb}{0.19, 0.55, 0.91}
\definecolor{blue(ncs)}{rgb}{0.265, 0.445, 0.765}
\definecolor{blue(ryb)}{rgb}{0.01, 0.28, 1.0}
\definecolor{orange}{rgb}{1.0, 0.49, 0.0}
\definecolor{Gray}{gray}{0.88}
\definecolor{green(ncs)}{rgb}{0.0, 0.62, 0.42}
\definecolor{clova}{rgb}{0.24, 0.63, 0.33}
\definecolor{correctgreen}{rgb}{0.01, 0.702, 0.321}
\definecolor{wrongred}{rgb}{0.757, 0, 0}

\definecolor{teaserblue}{rgb}{0.274, 0.694, 1}
\definecolor{teaserorange}{rgb}{0.913, 0.443, 0.196}
\definecolor{teaserred}{rgb}{1,0,0}

\definecolor{kellygreen}{rgb}{0.3, 0.73, 0.09}

\newcolumntype{g}{>{\columncolor{CuGray}}c}
\newcolumntype{z}{>{\columncolor{CuGray}}l}

\renewcommand{\paragraph}[1]{\noindent\textbf{#1.}\,\,}

\definecolor{steelblue}{rgb}{0.27, 0.51, 0.7}

\usepackage{xspace}

\makeatletter
\def\@fnsymbol#1{\ensuremath{\ifcase#1\or *\or \dagger\or \ddagger\or
   \mathsection\or \mathparagraph\or \|\or **\or \dagger\dagger
   \or \ddagger\ddagger \else\@ctrerr\fi}}
\makeatother

\def\onedot{.\@\xspace}

\def\etal{\emph{et al}\onedot}

\newcommand{\Sref}[1]{Sec.~\ref{#1}}

\newcommand{\Fref}[1]{Fig.~\ref{#1}}
\newcommand{\Tref}[1]{Table~\ref{#1}}

\newcommand{\be}{\begin{eqnarray}}
\newcommand{\ee}{\end{eqnarray}}
\newcommand{\bee}{\begin{eqnarray*}}
\newcommand{\eee}{\end{eqnarray*}}

\newcommand{\matrixb}{\left[ \begin{array}}
\newcommand{\matrixe}{\end{array} \right]}

\newcommand{\argmin}{\operatornamewithlimits{\arg \min}}



%% file: sec/0_abstract.tex
\begin{abstract}
We present VoiceCraft-Dub, a novel approach for automated video dubbing that synthesizes high-quality speech from text and facial cues. This task has broad applications in filmmaking, multimedia creation, and assisting voice-impaired individuals. Building on the success of Neural Codec Language Models (NCLMs) for speech synthesis, our method extends their capabilities by incorporating video features, ensuring that synthesized speech is time-synchronized and expressively aligned with facial movements while preserving natural prosody. To inject visual cues, we design adapters to align facial features with the NCLM token space and introduce audio-visual fusion layers to merge audio-visual information within the NCLM framework. Additionally, we curate CelebV-Dub, a new dataset of expressive, real-world videos specifically designed for automated video dubbing. Extensive experiments show that our model achieves high-quality, intelligible, and natural speech synthesis with accurate lip synchronization, outperforming existing methods in human perception and performing favorably in objective evaluations. We also adapt VoiceCraft-Dub for the video-to-speech task, demonstrating its versatility for various applications.
\end{abstract}

%% file: sec/1_intro.tex
\section{Introduction}\label{sec:intro}
Recent advancements in Text-to-Speech (TTS) have significantly improved the ability to synthesize human-like speech with remarkable naturalness, content accuracy, and voice similarity. Specifically, Neural Codec Language Models (NCLMs) represent a new model for speech generation that leverages language modeling with discrete codes derived from neural speech codecs~\cite{zeghidour2021soundstream, defossez2022high}. This approach has demonstrated superior performance across diverse speech generation 
tasks, such as TTS~\cite{valle1,valle2, kharitonov2023speak}, speech editing~\cite{peng2024voicecraft}, and voice conversion~\cite{baade2024neural}, due to the strong in-context learning ability of language models. Here, context refers to the speaker characteristics in the reference speech, enabling the generated speech to exhibit higher speaker similarity and naturalness compared to conventional models.

While TTS traditionally synthesizes speech from input text, combining text with video input gives rise to a different application: \textbf{automated video dubbing}~\cite{hu2021neural}. Automated video dubbing involves generating target speech given the source speech, target text, and target video, as shown in \Fref{fig:teaser} (a). The source speech provides reference voice characteristics, while the target text and video determine the content to be synthesized, its temporal alignment to the speaker's facial and lip movements, and even provide prosodic and emotional cues for how the speech should be delivered. 
This task has applications in filmmaking, multimedia content creation, dubbing translations of films into different languages, redubbing silent movies, and voice generation for voice-impaired individuals~\cite{nakamura2012speaking,kain2007improving}. 
Automated video dubbing is challenging as it must meet criteria inherited from TTS: (1) Intelligibility: the speech should convey accurate content, (2) Naturalness: the speech should exhibit natural prosody and intonation, and (3) Speaker similarity: the voice should match the source speaker. Additionally, video-specific criteria must also be considered: (4) Lip synchronization: the synthesized speech should be precisely time-aligned with lip movements, and (5) Expressiveness: the speech should reflect natural expressions corresponding to facial cues.

In this work, we introduce \textbf{VoiceCraft-Dub}, a novel approach for automated video dubbing using NCLMs. Building on the strengths of NCLMs in TTS, we extend their capabilities to incorporate video features, enabling the synthesis of natural and expressive speech that is time-synchronized with the target video. Specifically, we frame the task as an autoregressive speech token prediction problem, conditioned on the source speech, target text, and target video. Since NCLMs typically only take as input text and speech codec tokens, we introduce an audio-visual fusion approach to inform the model of the desired alignment between the output speech and facial features—such as lip movements and facial expressions—by merging the speech and visual tokens at each timestep. These merged audio-visual tokens are autoregressively fed into a Transformer decoder to generate the next speech token, enabling the model to synthesize time-aligned and expressive speech by referencing the target video, as shown in \Fref{fig:teaser} (b).

We evaluate our proposed method using the existing audio-visual dataset LRS3~\cite{afouras2018lrs3}, which is effective for assessing model performance in real-world scenarios. To further test our model's ability to generate expressive speech in diverse, in-the-wild settings, we curate a video dataset, \textbf{CelebV-Dub}, built upon existing video sources~\cite{celebvhq,celebvtext} that include emotionally rich scenarios from vlogs, dramas, and movies. To ensure dataset quality, we design a curation pipeline that filters content suitable for automated video dubbing. Compared to LRS3, CelebV-Dub includes a broader range of in-the-wild videos with expressive speech.

Given the generative nature of this task, subjective evaluation is essential for assessing the quality of synthesized speech in terms of naturalness and lip synchronization. We conduct extensive human evaluations where participants assess synthesized outputs from different approaches across various aspects, including naturalness, intelligibility, lip-sync accuracy, speaker similarity, and expressiveness. Additionally, we use several objective metrics for quantitative evaluation. Experimental results show that our method achieves highly accurate lip synchronization while significantly outperforming state-of-the-art approaches~\cite{hpmdubbing,styledubber} in naturalness and content accuracy across all metrics. Furthermore, we demonstrate the versatility of our approach by adapting it for the video-to-speech generation task using an off-the-shelf visual speech recognition model~\cite{vsr}. Although this is not our primary focus, our model performs favorably, emphasizing its potential for a wide range of applications. Our main contributions are summarized as follows:

\begin{itemize}
\item We introduce VoiceCraft-Dub, the first work to extend Neural Codec Language Models (NCLMs) for automated video dubbing by integrating visual facial cues.
\item We propose a novel audio-visual fusion approach in NCLMs to achieve precise lip synchronization while preserving speech quality and intelligibility.
\item We curate CelebV-Dub, a dataset containing expressive real-world videos designed for automated video dubbing.
\item We demonstrate superior performance in synthesizing human-like speech with high content accuracy, naturalness, and precise synchronization with video.
\end{itemize}

%% file: sec/2_related.tex
\section{Related work} \label{sec:related}
\paragraph{Neural codec language models (NCLMs)}
NCLMs are inspired by advancements in both neural audio codecs and language modeling techniques. Neural audio codecs~\cite{defossez2022high, zeghidour2021soundstream} have enabled high-fidelity audio compression by efficiently encoding raw waveforms into discrete tokens while preserving quality using residual vector quantization (RVQ) method. By applying language modeling techniques to these discrete audio tokens, NCLMs have been shown to generate high-quality speech. NCLMs have also demonstrated the ability to perform in-context learning by copying the vocal characteristics, emotion, and prosody of a prompt utterance, resulting in highly natural and expressive speech. In this regard, NCLMs significantly outperform conventional TTS methods. 
NCLMs have demonstrated strong performance in diverse applications, such as speech continuation~\cite{borsos2023audiolm}, zero-shot TTS~\cite{kharitonov2023speak, valle1, valle2}, speech editing~\cite{peng2024voicecraft}, and voice conversion~\cite{baade2024neural}. Several methods have also explored style-controlled synthesis~\cite{yang2024instructtts,liu2023promptstyle}. Furthermore, NCLMs have been successfully applied to other audio domains, such as music~\cite{agostinelli2023musiclm,garcia2023vampnet,copet2024simple} and sound effects~\cite{kreuk2022audiogen}.
We further extend the capabilities of NCLMs by integrating them with visual information, specifically targeting the automated video dubbing task. By bridging the audio-visual domain, we significantly expand the potential and versatility of NCLMs.

\begin{figure*}[t]
    \centering
    \small
    \includegraphics[width=1.0\linewidth]{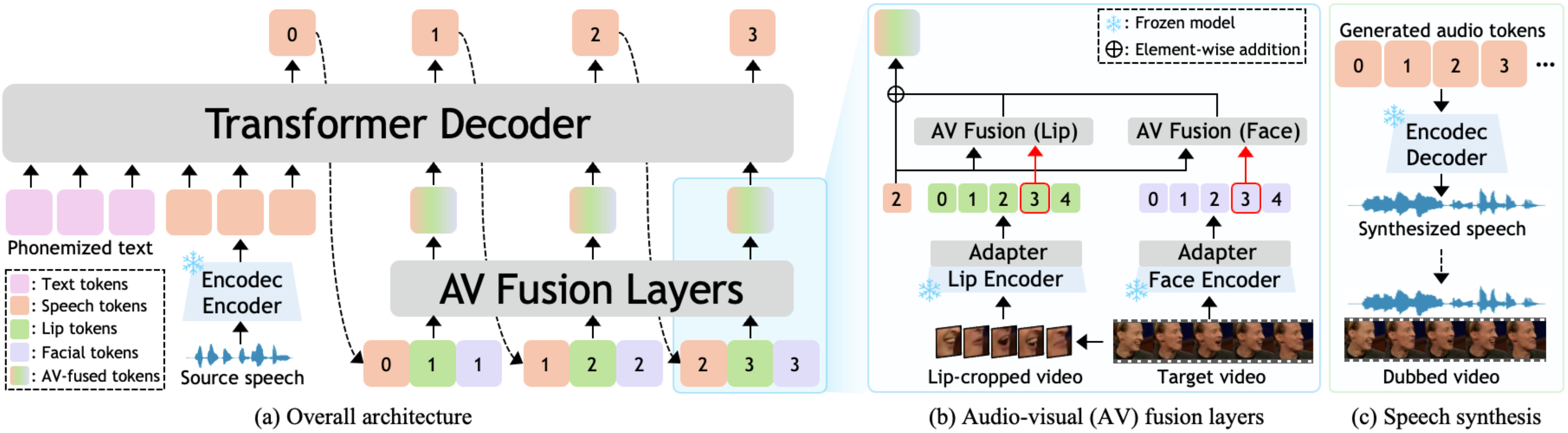}
    \vspace{-6mm}
    \caption{\textbf{Our proposed approach.} (a) The Transformer decoder autoregressively generates audio tokens from phonemized text tokens, source speech tokens (extracted via the Encodec encoder), and audio-visual fused tokens, which combine the target speech token with lip and facial tokens from the target video. The numbers in each token denote the timestep. (b) An audio-visual fusion layer aligns the generated target speech tokens with the preceding lip and face tokens, effectively merging their information for better synchronization. (c) Finally, the generated target speech tokens are decoded by Encodec to synthesize high-quality speech that is temporally aligned with the video.} 
    \label{fig:method}
    \vspace{-1mm}
\end{figure*}

\paragraph{Automated video dubbing}
Automated video dubbing is a task that synthesizes speech from both text and video inputs, with promising applications in multimedia creation. Since this task extends text-to-speech (TTS) by incorporating video, research in this field has largely evolved from advancements in TTS. Early work~\cite{hu2021neural,lu2022visualtts} primarily focuses on modifying TTS models to generate lip-synced speech using lip-cropped videos. Subsequent work~\cite{hassid2022more} proposes reflecting both text and the speaker's emotional state by analyzing facial expressions for speech synthesis.
Building on these foundations, HPMDubbing~\cite{hpmdubbing} introduces a multi-level approach that aligns visual cues with speech prosody across lip movements, facial expressions, and scene context, modifying FastSpeech~\cite{ren2019fastspeech} to incorporate visual cues. StyleDubber~\cite{styledubber} implements a multi-scale style learning framework to improve pronunciation accuracy while maintaining consistent speech style. Zhang~\etal~\cite{zhang2024speaker} propose a two-stage method: first, learning pronunciation from a large-scale text-to-speech corpus, then synchronizing prosody with speech and visual emotions, while ensuring duration consistency between speech timing and lip movements.

Despite these efforts, existing methods still struggle to synthesize natural and expressive human-like speech. To overcome these limitations, we bring NCLMs to bear for automated video dubbing, leveraging the strong in-context learning ability of these models for natural and high-fidelity speech synthesis. However, directly applying NCLMs off-the-shelf to this task is not possible, as existing NCLMs do not take video inputs as features, nor do they have mechanisms that can be used to precisely align the generated speech waveform with a target video of a talker's face. To address this, we propose a novel method for effectively injecting video features into NCLMs, enabling the synthesis of high-fidelity speech that aligns accurately with the video's timing and expressions.

%% file: sec/3_method.tex
\section{Modeling approach}\label{sec:method}
\subsection{Overview}
Our primary goal is to generate high-quality, human-like speech while ensuring precise temporal alignment with the target video. Given the source speech $\mathbf{Y}_{\text{src}}$, text input $\mathbf{Z}$, and target video $\mathbf{V}$, we design a model that synthesizes the target speech $\mathbf{Y}_{\text{tgt}}$. To achieve this, we formulate the task as a next-token prediction problem using a Neural Codec Language Model (NCLM), which has been shown to produce natural and expressive speech~\cite{borsos2023audiolm, kharitonov2023speak, valle1, valle2, peng2024voicecraft, baade2024neural}.

In conventional NCLM-based speech synthesis, the Transformer decoder processes either text alone or a combination of text and speech tokens to autoregressively generate speech. A straightforward extension for video dubbing is to prepend video tokens alongside text and speech tokens in the Transformer decoder. However, we empirically found that this increases the sequence length to the point that it becomes difficult for the model to effectively attend to each modality. This leads to performance degradation in content accuracy, intelligibility, and lip synchronization (refer to \Sref{sec:ablation}).

To address this issue, we propose VoiceCraft-Dub, a novel approach that extends the NCLM framework to incorporate visual cues for automated video dubbing. As shown in \Fref{fig:method} (a), VoiceCraft-Dub takes text and discrete source speech tokens as inputs and autoregressively generates target speech tokens, with each token merged with visual features through audio-visual (AV) fusion layers. Specifically, as described in \Fref{fig:method} (b), visual features, including lip movements and facial expressions, are mapped into the NCLM token space using separate adapters. These mapped tokens are fused with target speech tokens via AV fusion layers. Unlike simply prepending video tokens, this fusion mechanism directly aligns visual features with speech generation, enabling the Transformer decoder to produce speech that is temporally aligned with the video while maintaining intelligibility. Finally, the generated target speech tokens are fed into the neural codec decoder to synthesize high-quality speech, as in \Fref{fig:method} (c). In the following subsections, we detail the input features, architecture, and techniques used to achieve high-quality automated video dubbing.

\subsection{Input feature extraction}\label{sec:intput}
We describe the input representations for each modality that are provided to the Transformer decoder as inputs.

\paragraph{Speech input}
The source speech $\mathbf{Y}_\text{src}$ provides the speaker's voice and prosodic characteristics. We use discrete neural speech tokens extracted from the pre-trained Encodec~\cite{defossez2022high} as the speech input for the Transformer decoder. Encodec is a high-fidelity neural audio codec designed for efficient speech and audio compression. It encodes raw waveforms into a compact discrete representation using a residual vector quantization (RVQ) mechanism while preserving audio quality. The model consists of an encoder that transforms audio into quantized tokens, a decoder that reconstructs the waveform from these tokens, and codebooks that store learned representations for efficient compression.

Given the source speech $\mathbf{Y}_\text{src}$, we apply the Encodec encoder $E$ to extract a sequence of discrete speech tokens:
$\mathbf{\tilde{h}}_\text{src} = E(\mathbf{Y}_\text{src})$, where $\mathbf{\tilde{h}}_\text{src} \in \mathbb{R}^{T_{\text{src}} \times D \times K}$, $T_{\text{src}}$ is the number of timesteps, $D$ is the dimension of each token, and $K$ is the number of RVQ codebooks.
The Encodec model we use operates at a codec framerate of 50Hz on 16kHz recordings.

\paragraph{Text input}
The text input $\mathbf{Z}$ specifies the content to be synthesized.
To process text effectively, we convert it into phonemes based on the International Phonetic Alphabet (IPA) using the Phonemizer toolkit~\cite{bernard2021phonemizer}.
Each phoneme is mapped to a learnable vector representation, $\mathbf{h}_\text{text} \in \mathbb{R}^{T_{\text{text}} \times D}$, where $T_{\text{text}}$ is the length of the phoneme sequence and $D$ matches the dimension of the discrete speech tokens. These learnable representations are updated during model training.

\paragraph{Video input}
The target video $\mathbf{V}$ provides the necessary facial cues for speech synthesis. Specifically, the video input serves two primary roles: the full-face video ($\mathbf{V}$) captures facial expression context to enhance expressiveness, while the lip video ($\mathbf{V}_\text{lip}$), cropped from $\mathbf{V}$, indicates when to speak or pause, ensuring temporal synchronization.

We leverage off-the-shelf models to extract facial features: AV-HuBERT~\cite{avhubert}, a powerful audio-visual representation model, serves as a lip encoder to extract lip features from $\mathbf{V}_\text{lip}$, and EmoFAN~\cite{emofan}, a facial expression analysis model, serves as a face encoder to extract overall facial features from $\mathbf{V}$. Since these extracted features exist in different representational spaces compared to speech inputs, we introduce adapters composed of MLP layers to align them with NCLM’s token space, as shown in \Fref{fig:method} (b):

\begin{equation}
\textbf{h}'_\text{lip} = \text{Adapter}_\text{lip}(\text{AV-HuBERT}(\mathbf{V}_\text{lip}))
\end{equation}
\begin{equation}
\textbf{h}'_\text{face} = \text{Adapter}_\text{face}(\text{EmoFAN}(\mathbf{V}))
\end{equation}
where $\textbf{h}'_\text{lip}, \textbf{h}'_\text{face} \in \mathbb{R}^{M \times D}$, $M$ is the number of video frames and $D$ matches the token dimension. We further duplicate the video features to match the temporal resolution of speech tokens: $\mathbf{h}_\text{lip}, \mathbf{h}_\text{face} \in \mathbb{R}^{T \times D}$ where $T = 2M$, since video frames are at 25fps while speech tokens are at 50Hz.

\subsection{Audio-visual (AV) fusion layers}
Since extending NCLMs to maintain temporal synchronization between synthesized speech and video is non-trivial, our AV fusion layers play a crucial role by effectively integrating visual and speech information, enabling accurate speech generation synchronized with video.
As shown in \Fref{fig:method} (a), the AV fusion layers take speech, lip, and facial tokens at each timestep and merge them into AV-fused tokens, which are then provided to the Transformer decoder. We design two AV fusion layers, each combining speech tokens with lip and facial tokens, as described in \Fref{fig:method} (b). Specifically, given the currently generated target speech token $\mathbf{h}^t_\text{tgt}$, we fuse it with the lip and facial tokens as follows:
\begin{equation}
\mathbf{r}^t_\text{tgt, lip} = \text{AVFuse}_\text{lip}([\mathbf{h}^t_\text{tgt}; \mathbf{h}^{t+1}_\text{lip}])
\end{equation}
\begin{equation}
\mathbf{r}^t_\text{tgt, face} = \text{AVFuse}_\text{face}([\mathbf{h}^t_\text{tgt}; \mathbf{h}^{t+1}_\text{face}]),
\end{equation}
where $\mathbf{h}^t_\text{tgt}, \mathbf{r}^{t}_\text{tgt, lip}, \mathbf{r}^{t}_\text{tgt, face} \in \mathbb{R}^{D}$, $\text{AVFuse}_{\text{lip/face}}$ are the AV fusion layers, $t$ is the current timestep, and $[;]$ denotes channel-wise concatenation.
Importantly, we fuse the currently generated speech token $\mathbf{h}^t_\text{tgt}$ with the preceding lip and facial tokens at timestep $t+1$.
This allows the model to preview the upcoming visual changes to help the model synthesize temporally and semantically aligned next speech token with next frame video.
Finally, $\mathbf{r}^{t}_\text{tgt, lip}$ and $\mathbf{r}^{t}_\text{tgt, face}$ act as residual values, which are added to the current target token as $\mathbf{h}^t_\text{fuse} = \mathbf{h}^t_\text{tgt} + \mathbf{r}^t_\text{tgt, lip} + \mathbf{r}^t_\text{tgt, face}$ and then fed into the Transformer decoder for autoregressive speech generation.

\subsection{Transformer decoder}
The Transformer decoder follows a GPT-style architecture, with the goal of generating target speech tokens $\mathbf{\tilde{h}}_\text{tgt}$ up to $T$, aligned with the fixed-length video.
Specifically, the decoder synthesizes the current target speech token as
$\mathbf{h}_\text{tgt}^{t}=G([\mathbf{h}_\text{text}; \mathbf{h}_\text{src}; \mathbf{h}_\text{fuse}^{<t}])$, where $G$ is the Transformer decoder.
As mentioned in \Sref{sec:intput}, the source speech tokens $\mathbf{\tilde{h}}_\text{src} \in \mathbb{R}^{T_{\text{src}} \times D \times K}$ consist of $K$ residual codebooks. These tokens are summed along the codebook dimension, resulting in $\mathbf{h}_\text{src} \in \mathbb{R}^{T_{\text{src}} \times D}$ to serve as inputs for the decoder.
The generated target speech token $\mathbf{h}_\text{tgt}^t$ is passed through $K$ MLP heads to predict logits for each residual codebook, yielding $\mathbf{\tilde{h}}_\text{tgt}^t \in \mathbb{R}^{D \times K}$. This is summed along the codebook dimension before being passed into the audio-visual fusion layer in the next timestep. 
The text and source speech tokens are combined with their corresponding sinusoidal positional encoding~\cite{vaswani2017attention} before being fed into the Transformer decoder, while the AV-fused tokens share the positional encoding of the source speech.
Following existing approaches~\cite{peng2024voicecraft,copet2023simple}, we adopt a delayed prediction strategy, effective for autoregressive generation over stacked RVQ tokens. At each timestep $t$, each residual level is delayed, so the prediction of codebook $k-1$ at timestep $t$ is conditioned by the prediction of codebook $k$ at the same timestep.

Once the generation is complete, the discrete speech tokens are passed to the Encodec decoder $D$: $\mathbf{Y}_\text{tgt} = D(\mathbf{\tilde{h}}_\text{tgt})$ to reconstruct the waveform, as shown in \Fref{fig:method} (c).

\paragraph{Learning objective}
The model is trained to minimize the negative log-likelihood of predicting the correct speech tokens, given the text, source speech, and fused audio-visual tokens. Specifically, following the insights in \cite{peng2024voicecraft} that weighting the first residual codebook more for updates is effective, we assign higher weights to the first residual codebook compared to the later ones, resulting in the final loss:
\begin{equation}
L(\theta) = -\sum_{k=1}^{K} \alpha_k \log P_\theta(\mathbf{\tilde{h}}_{\text{tgt},k} | \mathbf{h}_\text{src}, \mathbf{h}_\text{text}, \mathbf{h}_\text{fuse}),
\end{equation}
where $\alpha_k$ denotes the weighting hyperparameters. The trainable parameters include the adapters, audio-visual fusion layers, and the Transformer decoder, while AV-HuBERT, EmoFAN, and Encodec remain frozen.

\subsection{Implementation details}
The Transformer decoder is initialized with pretrained weights from VoiceCraft~\cite{peng2024voicecraft} and fine-tuned, while the audio-visual fusion layers and adapters are trained from scratch.
Encodec has $K=4$ RVQ codebooks, each with a vocabulary size of $D=2048$. The Transformer decoder consists of 16 layers, with hidden and FFN dimensions of 2048 and 8192, and 12 attention heads. The weight hyperparameters for updating the codebook are set to $\alpha=[3,1,1,1]$. The audio-visual fusion layers each contain a linear layer, and the adapters are designed as two-layer MLPs with GELU activation~\cite{gelu}.
For training, we use the AdamW optimizer with a base learning rate of $1\text{e-}5$ for the Transformer decoder and $1\text{e-}2$ for the other modules. We train using four RTX 8000 GPUs, with a total batch size of 80K frames, and conduct training for 100K steps with early stopping.

%% file: sec/4_exp.tex
\section{Experiments}\label{sec:exp}
We validate the effectiveness of our proposed model through a series of subjective and objective assessments. We further provide ablations of our design choices and demonstrate the extended application of our approach in video-to-speech generation.

\subsection{Experimental setup}\label{sec:exp_setup}
\paragraph{Dataset}
We train and test our model on the LRS3 dataset~\cite{afouras2018lrs3} and our curated CelebV-Dub dataset. LRS3 is an in-the-wild video dataset in English, sourced from TED and TEDx talks. It comprises approximately 439 hours of video with unconstrained, long utterances from thousands of speakers. The pretrain split is utilized for training, while 1,174 utterances from the test split are reserved for testing. 

Since LRS3 videos predominantly feature neutral expressions, we introduce CelebV-Dub to enhance our evaluation. This dataset is built upon CelebV-HQ~\cite{celebvhq} and CelebV-Text~\cite{celebvtext}, collected from diverse in-the-wild sources such as vlogs, dramas, and movies. CelebV-Dub serves as a challenging benchmark by featuring unconstrained real-world settings and expressive variations. Since CelebV-HQ and CelebV-Text contain considerable noise, we develop a dataset curation algorithm to extract expressive in-the-wild videos featuring active speakers, segmented into utterances with minimal background noise, and labeled with pseudo-transcriptions. More details about the dataset construction are provided in the Sec.~\textcolor{iccvblue}{B} of the supplementary material.

\begin{table*}[tp]
    \centering
    \resizebox{0.9\linewidth}{!}{\begin{tabular}{lccccccccccc}
    \toprule
    \multirow{2}{*}[-0.4em]{Model} & \multicolumn{2}{c}{Naturalness} & \multicolumn{2}{c}{Intelligibility}& \multicolumn{2}{c}{Lip synchronization}& \multicolumn{2}{c}{Expressiveness} & \multicolumn{2}{c}{Speaker similarity} \\
    \cmidrule(r{2mm}l{2mm}){2-3} \cmidrule(r{2mm}l{2mm}){4-5} \cmidrule(r{2mm}l{2mm}){6-7} \cmidrule(r{2mm}l{2mm}){8-9} \cmidrule(r{2mm}l{2mm}){10-11}
    & LRS3 & CelebV-Dub & LRS3 & CelebV-Dub & LRS3 & CelebV-Dub & LRS3 & CelebV-Dub& LRS3 & CelebV-Dub \\ 
    \cmidrule{1-11}
    Ground-Truth &4.42{\tiny$\pm$0.07} & 4.42{\tiny$\pm$0.11} &  4.65{\tiny$\pm$0.05} & 4.58{\tiny$\pm$0.08} & 4.45{\tiny$\pm$0.06} & 4.50{\tiny$\pm$0.11}& 4.38{\tiny$\pm$0.07}  & 4.52{\tiny$\pm$0.09} & 3.56{\tiny$\pm$0.06} & 3.58{\tiny$\pm$0.09} \\
    \cmidrule{1-11}
    Ours &\textbf{4.30{\tiny$\pm$0.07}}  & \textbf{4.18{\tiny$\pm$0.11}} & \textbf{4.52{\tiny$\pm$0.06}}  & \textbf{4.42{\tiny$\pm$0.10}} & \textbf{4.37{\tiny$\pm$0.06}} & \textbf{4.42{\tiny$\pm$0.11}}& \textbf{4.33{\tiny$\pm$0.07}} &\textbf{ 4.41{\tiny$\pm$0.09}} & \textbf{3.36{\tiny$\pm$0.07}}& \textbf{3.44{\tiny$\pm$0.09}} \\
    HPMDubbing~\cite{hpmdubbing} & 3.12{\tiny$\pm$0.11} & 1.65{\tiny$\pm$0.09} & 3.33{\tiny$\pm$0.09} & 1.74{\tiny$\pm$0.10} & 3.96{\tiny$\pm$0.07} & 3.34{\tiny$\pm$0.15}& 2.93{\tiny$\pm$0.10}& 2.34{\tiny$\pm$0.14} & 1.66{\tiny$\pm$0.09} & 0.72{\tiny$\pm$0.10} \\ 
    StyleDubber~\cite{styledubber} &2.85{\tiny$\pm$0.11} & 1.68{\tiny$\pm$0.09} & 3.40{\tiny$\pm$0.09}  & 2.24{\tiny$\pm$0.13} & 3.83{\tiny$\pm$0.08}  & 3.14{\tiny$\pm$0.15} & 2.78{\tiny$\pm$0.10} & 2.26{\tiny$\pm$0.13} & 1.50{\tiny$\pm$0.10} & 0.92{\tiny$\pm$0.11} \\ 
    \bottomrule
    \end{tabular}}
    \vspace{-1mm}
    \caption{\textbf{Comparison of Mean Opinion Score (MOS).} Our proposed VoiceCraft-Dub significantly outperforms existing methods and performs on par with the ground truth in diverse aspects of human perception metrics.}
    \label{tab:mos}
\end{table*}

\begin{table*}[tp]
    \centering
    \resizebox{0.9\linewidth}{!}{\begin{tabular}{lccccccccc}
    \toprule
    \multirow{2}{*}[-0.4em]{A vs. B} & \multicolumn{3}{c}{Naturalness} & \multicolumn{3}{c}{Expressiveness}& \multicolumn{3}{c}{Lip synchronization}\\
    \cmidrule(r{2mm}l{2mm}){2-4} \cmidrule(r{2mm}l{2mm}){5-7} \cmidrule(r{2mm}l{2mm}){8-10}
    & A wins (\%) & Neutral & B wins (\%) & A wins (\%) & Neutral & B wins (\%) & A wins (\%) & Neutral & B wins (\%) \\ 
    \cmidrule{1-10}
    Ours vs. HPMDubbing~\cite{hpmdubbing} & \textbf{87.4} & 2.2 & 10.4 & \textbf{88.6} & 2.4 & 9.0 & \textbf{75.4} & 4.8 & 19.8\\
    Ours vs. StyleDubber~\cite{styledubber} & \textbf{96.6} & 0.6 & 2.8 & \textbf{97.4} & 0.4 & 2.2 & \textbf{85.3} & 6.6 & 8.1\\
    \cmidrule{1-10}
    Ground-Truth vs. HPMDubbing~\cite{hpmdubbing} & 95.0 & 1.6 & 3.4 & 95.8 & 1.8& 2.4 & 83.6 & 7.1 & 9.3\\
    Ground-Truth vs. Ours & \cellcolor{gray!20} 55.8 & 30.2& 14.0 & \cellcolor{gray!20} 56.0 & 30.4 & 13.6 & \cellcolor{gray!20} 61.0 & 14.8 & 24.2\\
    \bottomrule
    \end{tabular}}
    \vspace{-1mm}
    \caption{\textbf{A/B testing results on LRS3.} We report the preferences (\%) between A and B across various aspects of synthesized speech. In rows 1 and 2, our model is significantly preferred by humans over existing methods. Comparing rows 3 and 4, HPMDubbing is significantly less preferred compared to the ground truth, while our model, highlighted with a \colorbox{gray!20}{gray background}, is preferred more.}
    \label{tab:ab}
\end{table*}

\paragraph{Subjective metrics}
Given the generative nature of the task, subjective evaluation is essential to assess the quality of synthesized speech. We conduct two types of evaluations: Mean Opinion Score (MOS) and A/B testing. In the MOS evaluation, participants assess each synthesized speech for naturalness, lip-sync accuracy, intelligibility, expressiveness, and speaker similarity. Each sample is presented alongside the target video, and participants rate it on a 5-point Likert scale, where 1 indicates poor quality and 5 indicates excellent quality. In the A/B testing, we compare the output of VoiceCraft-Dub with existing methods, asking participants to choose which utterance sounds preferable in terms of naturalness, expressiveness, and lip-sync. These evaluations are conducted on 150 utterance samples (100 from LRS3 and 50 from CelebV-Dub) using Amazon Mechanical Turk.

\paragraph{Objective metrics}
We employ several metrics to evaluate various aspects of synthesized speech. Following prior work~\cite{peng2024voicecraft,valle1}, Word Error Rate (WER) assesses content accuracy, while speaker similarity (spkSIM) measures voice consistency, computed using the Whisper~\cite{whisper} medium.en model and WavLM-TDNN~\cite{wavlm}, respectively. Lip-sync accuracy (LSE-D and LSE-C) is evaluated using SyncNet~\cite{syncnet}, which takes both speech and video as inputs. Emotion similarity (emoSIM) is assessed by measuring cosine similarity between synthesized and ground truth speech embeddings using Emotion2Vec~\cite{emotion2vec}. We utilize automatic Mean Opinion Scores (MOS), DNSMOS~\cite{dnsmos} and UTMOS~\cite{utmos}, to evaluate overall speech quality. Low-level metrics, including Mel-Cepstral Distortion (MCD), fundamental frequency distance (F0), and energy distance (Energy), are also employed.

\begin{table*}[tp]
    \centering
    \resizebox{0.95\linewidth}{!}{\begin{tabular}{lcccccccccc}
    \toprule
    Method &  WER ($\downarrow$)& LSE-D ($\downarrow$)& LSE-C ($\uparrow$) & spkSIM ($\uparrow$) & UTMOS ($\uparrow$) & DNSMOS ($\uparrow$) & MCD ($\downarrow$)& F0 ($\downarrow$) & Energy ($\downarrow$) & emoSIM ($\uparrow$)\\
    \cmidrule{1-11}
    Ground-Truth & 1.79 & 6.88 & 7.63 & - & 3.59 & 3.18 & - & - & - & -\\
    Zero-shot TTS~\cite{peng2024voicecraft} & 0.68 & 10.41 & 4.04 & 0.333 & 3.48 & 3.30 & - & - & - & 0.682\\
    \cmidrule{1-11}
    HPMDubbing~\cite{hpmdubbing}& 7.19 & \textbf{6.58} & \textbf{7.99} & 0.219 & 3.08 & 2.98 & 8.70 & 61.54 & 3.01 & 0.743\\ 
    StyleDubber~\cite{styledubber} &3.25 & 9.33 & 5.21 & 0.295 & 2.71 & 2.95& 8.29 & 112.80 & 2.31&\uline{0.779}\\ 
    Ours (lip-only) &\textbf{1.38} & \uline{6.59} & \uline{7.97} & \uline{0.361} & \uline{3.66} & \uline{3.20}& \uline{7.84} & \uline{57.87} & \uline{1.96} & \textbf{0.789}\\ 
    Ours (lip $\&$ face) &\uline{1.68} & 6.87 & 7.74 & \textbf{0.373} & \textbf{3.86} & \textbf{3.21}& \textbf{7.58} & \textbf{54.01} & \textbf{1.92} & \textbf{0.789}\\ 
    \bottomrule

    \end{tabular}}
    \vspace{-2mm}
    \caption{\textbf{Comparison on the LRS3 dataset.} We highlight the best results in \textbf{Bold} and \uline{underline} the second best.}
    \label{tab:lrs}
\end{table*}

\begin{table*}[tp]
    \centering
    \resizebox{0.95\linewidth}{!}{\begin{tabular}{lcccccccccc}
    \toprule
    Method &  WER ($\downarrow$)& LSE-D ($\downarrow$)& LSE-C ($\uparrow$) & spkSIM ($\uparrow$) & UTMOS ($\uparrow$) & DNSMOS ($\uparrow$) & MCD ($\downarrow$)& F0 ($\downarrow$) & Energy ($\downarrow$) & emoSIM ($\uparrow$)\\
    \cmidrule{1-11}
    Ground-Truth &4.15 & 7.44 & 6.73 & - & 2.90 & 3.38& - & - & - & -\\
    Zero-shot TTS~\cite{peng2024voicecraft} & 3.83 & 11.68 & 2.78 & 0.316 & 2.92 & 3.40 & - & - & - & 0.704\\
    \cmidrule{1-11}
    HPMDubbing~\cite{hpmdubbing} &24.06 & \textbf{7.80} & \textbf{6.36} & 0.146 & 2.10 & 2.87& 9.80 & 107.04 & 4.38 & 0.721\\  
    StyleDubber~\cite{styledubber} &9.48 & 10.40 & 3.78 & 0.264 & 1.86 & 2.94& 8.18 & 180.68 & \uline{3.11} & \textbf{0.784}\\ 
    Ours (lip-only) &\uline{7.26} & 8.27 & 5.93 & \textbf{0.340} & \textbf{3.39} & \textbf{3.40}& \uline{7.97} & \uline{75.88} & 3.21 & 0.778\\ 
    Ours (lip $\&$ face) &\textbf{7.01} & \uline{8.13} & \uline{6.05} & \uline{0.333} & \uline{3.37} & \textbf{3.40}& \textbf{7.91} & \textbf{72.71} & \textbf{3.03} & \uline{0.782}\\ 
    \bottomrule
    \end{tabular}}
    \vspace{-2mm}
    \caption{\textbf{Comparison on our curated CelebV-Dub dataset.} We highlight the best results in \textbf{Bold} and \uline{underline} the second best.}
    \label{tab:celeb}
\end{table*}

\begin{table*}[!htbp]
    \centering
    \resizebox{1\linewidth}{!}{\begin{tabular}{lccccccccccccc}
    \toprule
     & P.E.& AV fuse& Face &WER ($\downarrow$)& LSE-D ($\downarrow$)& LSE-C ($\uparrow$) & SIM ($\uparrow$) & UTMOS ($\uparrow$) & DNSMOS ($\uparrow$) & MCD ($\downarrow$)& F0 ($\downarrow$) & Energy ($\downarrow$) & Emotion ($\uparrow$)\\
    \cmidrule{1-14}
    (a) &&&&2.85 & 7.61 & 6.95 & 0.361 & 3.68 & 3.18& 7.98 & \textbf{53.02} & 2.11 &0.770\\
    (b) &\checkmark&&&2.81 & 7.58 & 6.96 & 0.360 & 3.66 & 3.15& 7.91 & 56.82 &2.08 &0.768\\
    (c) &\checkmark&\checkmark&&\textbf{1.38} & \textbf{6.59} & \textbf{7.97} & 0.361 & 3.66 & 3.20& 7.84 & 57.87 & 1.96 & \textbf{0.789}\\
    (d) &\checkmark&\checkmark&\checkmark&1.68 & 6.87 & 7.74 & \textbf{0.373} & \textbf{3.86} & \textbf{3.21}& \textbf{7.58} & 54.01 & \textbf{1.92} & \textbf{0.789}\\ 
    \bottomrule
    \end{tabular}}
    \vspace{-2mm}
    \caption{\textbf{Abaltion results on the LRS3 dataset~\cite{afouras2018lrs3}.} We evaluate the effectiveness of various design choices by comparing different configurations of our method. ``P.E.'' denotes shared positional encoding between speech and video tokens, ``AV fuse'' indicates using audio-visual fusion layers for merging speech and video tokens, and ``Face'' refers to using both facial and lip features as inputs. (c) and (d) are the final models used in all other experiments.}
    \vspace{-2mm}
    \label{tab:ablation}
\end{table*}

\paragraph{Competing methods}
We compare our method with two open-source approaches: HPMDubbing~\cite{hpmdubbing} and StyleDubber~\cite{styledubber}, which are non-NCLM-based models. Both models use a FastSpeech-like decoder~\cite{ren2019fastspeech} that relies on mel-spectrogram representations for speech synthesis and requires explicit speech-text alignment through a forced aligner during training. To ensure a fair comparison with our model and improve their generalization, we train HPMDubbing and StyleDubber using both LRS3 and our proposed CelebV-Dub, as the original models were trained on relatively small datasets. Both models are trained with a batch size of 16 for over 500k steps until convergence, with the rest of the settings following the original configurations~\cite{hpmdubbing, styledubber}.

\paragraph{Inference}
For each test sample, we form \{source speech, target text, target video\} triplets as input for model inference, where the source speech is from a different utterance of the target speaker. For inference, we employ Nucleus sampling~\cite{holtzman2019curious} with p=0.8 and a temperature of 1 for all experiments. Although the model produces natural and accurately lip-synced speech, the stochastic nature of autoregressive generation can sometimes result in inaccurate sounds. Thus, similar to the sorting method described in \cite{valle2}, we generate 10 samples, sort them by WER and LSE-D, and select the optimal speech from the sorted examples. We apply this same sorting method to other methods during testing.

\subsection{Human evaluation}\label{sec:human}
As mentioned in \Sref{sec:exp_setup}, validating our model through human perception is the most crucial metric for genuinely evaluating performance. 
Table~\ref{tab:mos} shows the Mean Opinion Score (MOS) of our model, along with existing work~\cite{hpmdubbing, styledubber} and the ground truth, on the LRS3 and our collected CelebV-Dub dataset. As shown, our model outperforms existing methods across a wide range of evaluated criteria. Surprisingly, our model performs very closely to the ground truth in many cases. While the existing methods demonstrate favorable results in lip synchronization accuracy, our model significantly excels in naturalness, expressiveness, and speaker similarity.

Furthermore, Table~\ref{tab:ab} reports the A/B testing results on LRS3, summarizing preferences between samples from two different methods. In the top two rows, our model is significantly preferred over existing methods, achieving over 88\% preference in naturalness and expressiveness, and 75.4\% in lip synchronization. Comparing rows 3 and 4, we observe that HPMDubbing is significantly less preferred than our model when compared to the ground truth. Interestingly, in row 4, over 40\% of the time, humans perceive our output as comparable to or better than the ground truth. After the human evaluation, several comments noted that \emph{the synthesized speech from VoiceCraft-Dub is natural and time-synced with the video, making it hard to distinguish from real recordings.} These results highlight that the synthesized speech from our model is human-like, natural, and expressive enough to outperform existing methods while competing in quality with the ground truth, emphasizing the effectiveness of our approach. The A/B testing results on CelebV-Dub are available in the Sec.~\textcolor{iccvblue}{D.1} of the supplementary materials.

\begin{figure*}[tp]
    \centering
    \small
    \includegraphics[width=1.0\linewidth]{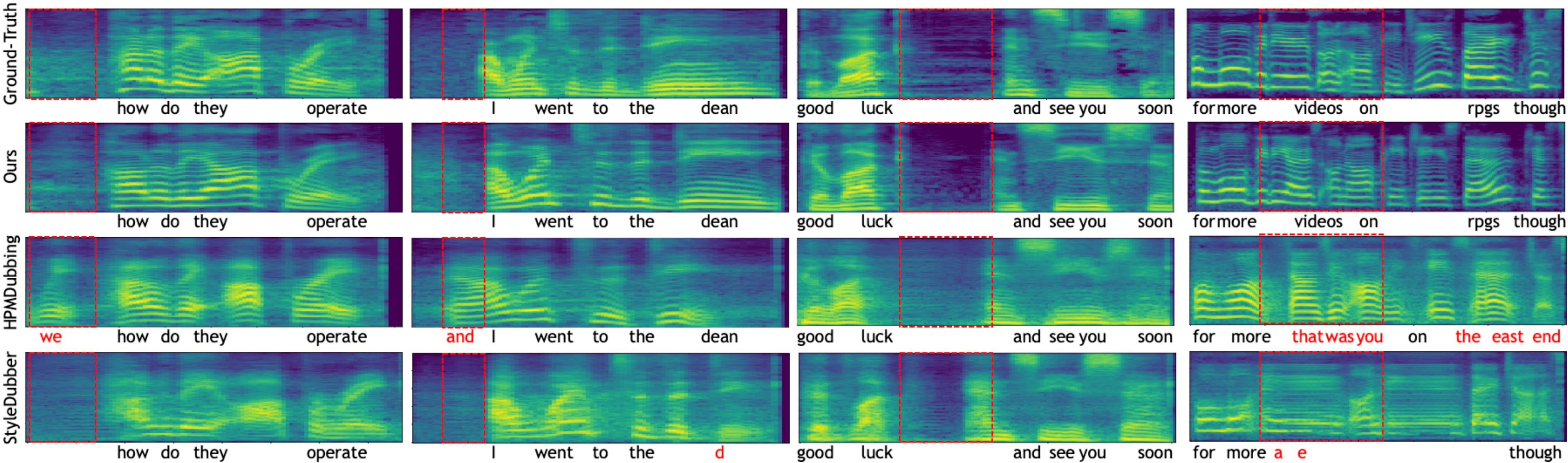}
    \vspace{-5mm}
    \caption{\textbf{Qualitative results.} We compare mel-spectrogram visualizations from ground truth recordings, our model, and prior methods~\cite{hpmdubbing, styledubber} on LRS3 (columns 1–2) and CelebV-Dub (columns 3–4). The texts below each mel-spectrogram represent time-aligned speech extracted using Whisper~\cite{whisper}, with red text indicating incorrectly synthesized speech.} 
    \vspace{-1mm}
    \label{fig:qual}
\end{figure*}

\subsection{Quantitative results}\label{sec:quan}
We provide quantitative comparisons on each dataset to assess various aspects of the synthesized speech. Specifically, we present results from two variants of our model: one that uses both lip and face input and another that uses only lip video input. In addition to the results from our models and existing methods, we also include the performance of the ground truth and an NCLM-based zero-shot TTS model~\cite{peng2024voicecraft} for reference. Note that the following results are intended to supplement the human evaluation; therefore, we recommend watching the demos for a more accurate assessment.

\Tref{tab:lrs} summarizes the comparison on LRS3. In terms of content accuracy (WER), speaker similarity (spkSIM), and overall speech quality (UTMOS, DNSMOS), our models outperform existing methods by a significant margin. Although HPMDubbing achieves the best lip-sync accuracy (LSE-D/C), our models perform on par with the ground truth. The zero-shot TTS model shows the best WER; however, this approach alone is unsuitable for video dubbing, as it lacks visual cues, leading to poor lip-sync accuracy and emotion similarity. Our results show that our approach effectively extends NCLM-based models to synthesize accurate lip-synced speech while maintaining intelligibility and naturalness.
When comparing our two models, we observe a slight degradation in WER and LSE-D/C with both lip and face input. Nevertheless, performance remains comparable to the ground truth and generally improves across other metrics. Since the LRS3 dataset mostly contains neutral expressions, no significant difference in emotion similarity (emoSIM) is observed between the two variants.

Similar trends are observed in comparisons conducted on our curated CelebV-Dub dataset, as summarized in \Tref{tab:celeb}. Again, the zero-shot TTS model achieves the best WER but exhibits poor lip-sync accuracy, highlighting the need of properly integrating visual cues. Both of our models successfully incorporate these cues and generally perform favorably against existing methods. One observation is that our models show lower LSE-D/C compared to HPMDubbing, which we attribute to the instability of SyncNet, a limitation noted in related works~\cite{yaman2024audio, ma2025sayanything, yaman2024cvpr}. We further provide an analysis in Sec.~\textcolor{iccvblue}{C} of the supplementary materials showing that LSE-D has a low correlation with human evaluation, suggesting it should be used as a reference rather than a definitive metric. Consequently, we argue that the human evaluation described in \Sref{sec:human} should be given greater weight as a more accurate measure. Since CelebV-Dub is a more expressive dataset compared to LRS3, we observe an improvement in emoSIM when using both lip and face input over the lip-only variant.

\subsection{Ablation study}\label{sec:ablation}
We conduct a series of experiments to verify our design choices, as detailed in \Tref{tab:ablation}. (a) extends the NCLM-based method by prepending video inputs, with video and speech inputs having separate positional encodings, while (b) is the same as (a) but with video tokens sharing the positional encoding of the speech. Comparing (a) and (b), we observe that using shared positional encoding does not significantly affect performance. Therefore, we introduce audio-visual fusion layers as in (c), which significantly improves WER and LSE-D/C compared to both (a) and (b). 
Using face input alongside lip input (d) shows comparable performance in WER and LSE-D/C to (c) and generally yields better performance. The impact of adding face input is further demonstrated in \Tref{tab:celeb}, and the generalization results in Sec.~\textcolor{iccvblue}{D.2} of the supplementary materials, particularly enhancing the emotion similarity (emoSIM). These insights led us to select (c) and (d) as the final model configurations.

\subsection{Qualitative results}
In \Fref{fig:qual}, we visually compare the mel-spectrogram samples converted from the synthesized speech of existing work~\cite{hpmdubbing, styledubber} and our VoiceCraft-Dub, along with those from ground truth recordings. Focusing on the red boxes in columns 1 and 2, we observe that HPMDubbing produces incorrect speech. Although StyleDubber synthesizes accurate content, its mel-spectrograms lack detail and appear blurry. These observations align with the results in \Sref{sec:human} and \Sref{sec:quan}, where StyleDubber received the lowest human and automatic MOS scores. In column 3, all models synthesize correct speech, but HPMDubbing and StyleDubber generate noticeable noise during brief pauses, as indicated by the red boxes. In column 4, existing methods produce incorrect content and blurry speech with considerable noise. In contrast, samples from our model across all experiments exhibit fine details in mel frequency and closely match the ground truth. These results demonstrate that our model synthesizes speech with accurate content and high quality.

\begin{table}[t]
    \centering
    \resizebox{1\linewidth}{!}{\begin{tabular}{@{}l@{}c@{\,\,\,}c@{\,\,\,}c@{\,\,\,}c@{\,\,\,}c@{}}
    
    \toprule
    Method & WER ($\downarrow$)& LSE-C ($\uparrow$) & spkSIM ($\uparrow$) & UTMOS ($\uparrow$) & DNSMOS ($\uparrow$)\\
    \cmidrule{1-6}
    VSR~\cite{vsr} & 26.75 & - & - & - & -\\ 
    VSR~\cite{vsr} \& TTS~\cite{peng2024voicecraft} & 29.72 & 3.66& 0.321 & 3.48 & 3.30\\ 
    \cmidrule{1-6}
    SVTS~\cite{mira2022svts}& 82.38 & 6.02 & 0.077 & 1.28 & 2.38\\ 
    Intelligible~\cite{choi2023intelligible}& \uline{30.00} &\textbf{8.02} & \uline{0.310} & \uline{2.70} & \uline{2.86}\\ 
    Ours (lip-only)& \textbf{28.83} &\uline{6.31}  &\textbf{0.334} & \textbf{3.52} & \textbf{3.16}\\ 
    \bottomrule
    \end{tabular}}
    \vspace{-2mm}
    \caption{\textbf{Results on video-to-speech generation on LRS3~\cite{afouras2018lrs3}.} Although not specifically designed for this task, our model performs favorably, particularly in content accuracy (WER) and overall quality (UTMOS and DNSMOS) compared to existing methods.}
    \vspace{-2mm}
    \label{tab:vsr}
\end{table}

\subsection{Application: Video-to-speech generation}
We demonstrate the extensibility of our model by adapting it to the video-to-speech synthesis task. Unlike automated video dubbing, video-to-speech generation requires only source speech and video inputs (without text) to synthesize speech. We use an expert Visual Speech Recognition (VSR) model~\cite{vsr} to extract text from the silent video. The extracted text, along with the provided video and source speech, is then fed into our model for video-to-speech generation.

\Tref{tab:vsr} compares the performance of our approach with several other methods, including standalone VSR, a combination of VSR and zero-shot TTS~\cite{peng2024voicecraft}, and the existing methods~\cite{vsr, mira2022svts}. Although video-to-speech synthesis is not our primary focus, the results show that our model performs favorably against the existing methods across various metrics. While the combination of VSR and TTS models achieves similar performance, as expected, it fails to synchronize with the video. This highlights that our approach effectively extends the capabilities of current TTS models to handle video inputs robustly. These results confirm that extending the NCLM-based model to accommodate visual inputs is both versatile and effective for speech synthesis.

%% file: sec/5_conclusion.tex
\section{Discussion and conclusion}\label{sec:conclusion}
\paragraph{Discussion} While our method shows high-quality speech synthesis, it also has limitations. Despite the effectiveness of full-face input, the model sometimes follows the characteristics of the source speech over the face input. Additionally, the autoregressive approach can produce incorrect samples if the next token prediction fails. Nonetheless, the results show that our approach is effective in synthesizing expressive and accurately time-aligned speech. Explicitly infusing emotional state or specifying speech length could address these limitations, which we plan to explore in future work.

\paragraph{Conclusion}
In this work, we propose VoiceCraft-Dub, a novel NCLM-based approach for automatic video dubbing. We extend the high-quality, human-like speech synthesis capabilities of NCLMs to incorporate visual cues as inputs. Specifically, we perform audio-visual fusion to provide direct alignment signals to NCLM for visually aligned speech synthesis. Additionally, we curate the expressive CelebV-Dub dataset, specifically designed for dubbing tasks. Our extensive human evaluations and quantitative results show that VoiceCraft-Dub outperforms existing methods and performs on par with the original recordings in terms of naturalness, intelligibility, and lip synchronization. We also demonstrate the versatility of our approach by adapting it to the video-to-speech generation task. We believe our approach will lead to a new paradigm in human-like video dubbing synthesis, enabling more immersive content creation.

%% file: supplement.tex
\renewcommand{\thefigure}{S\arabic{figure}}
\renewcommand{\thetable}{S\arabic{table}}
\setcounter{figure}{0} 
\setcounter{table}{0} 

In this supplementary material, we provide details on the dataset construction pipeline, additional results and analysis, implementation details, metrics, and human evaluations that were not included in the main paper.

\section{Dataset and code}
For reproducibility, we plan to open-source our training and inference code, along with the curated dataset and its annotations, upon acceptance.

\section{Data curation pipeline for CelebV-Dub}
We introduce the CelebV-Dub dataset, consisting of expressive video clips specifically suitable for automated video dubbing tasks. Despite the abundance of existing talking-video datasets~\cite{afouras2018lrs3, voxceleb2, wang2020mead, sung2024multitalk}, our goal is to curate in-the-wild videos that capture natural yet expressive speech. Such videos are effective for training and testing automated video dubbing models, which require synthesizing not only neutral but also expressive speech synchronized with facial cues. Our curated dataset comprises multiple speakers and utterances, each accompanied by a corresponding transcript. The dataset statistics are summarized in \Tref{tab:dataset_statistics}.

\paragraph{Video collection}
We initially collect videos from existing sources, including CelebV-HQ~\cite{celebvhq} and CelebV-Text~\cite{celebvtext}. These datasets originate from diverse sources, such as vlogs, dramas, and influencer videos, providing expressive, in-the-wild content across various scenarios. However, the provided metadata from these datasets varies in length—from single utterances to longer sequences—and contains substantial noise, such as non-active speakers and occluded faces. Therefore, we design a data curation pipeline to collect suitable videos specifically for automated video dubbing.

\paragraph{Detecting English and labeling pseudo-transcripts}
For each video in the existing sources, we use WhisperX~\cite{bain2023whisperx} to detect the language and generate pseudo-transcripts automatically. In constructing this dataset, we retain only videos identified as containing English speech, discarding all others.

\paragraph{Trimming videos into utterances}
WhisperX provides timestamps at the word and sentence levels, enabling precise video segmentation. Given the variability in utterance lengths of the original videos, we unify the dataset by trimming each video clip to contain a single utterance, utilizing the timestamps provided by WhisperX.

\paragraph{Frontal face verification}
The trimmed videos occasionally contain faces that are not oriented toward the front, preventing the models from learning distinct facial movements corresponding to speech. To address this, we measure yaw and pitch angles using Mediapipe~\cite{mediapipe} and remove clips with abrupt head movements or large yaw and pitch angles, which indicate side-facing poses.

\paragraph{Active speaker detection}
Training videos for automated video dubbing require facial movements synchronized with speech. To ensure this synchronization, we apply TalkNet~\cite{tao2021someone}, a model that employs audio-visual cross-attention to identify active speakers. We set conservative thresholds to minimize false positives, ensuring that only videos clearly containing active speakers are retained. Clips that do not meet these thresholds are discarded.

\paragraph{Background music suppression}
Background music in audio tracks can disturb clear speech signals necessary for effective model training. We employ Spleeter~\cite{spleeter2020} to detect and suppress background music where present, thereby preserving the clarity of speech signals.

\begin{table}
  \centering
  \resizebox{0.93\linewidth}{!}{
    \begin{tabular}{lccc}
    \toprule
    & Train & Test & Total\\
    \midrule
    Number of total video clips & 67,549 & 216& 67,765\\
    Number of speakers & 6,530 & 33 &6,563\\
    Average utterances per speaker & 10.34& 6.55& 10.33\\ 
    Average duration (seconds) &4.58 & 3.39& 4.57\\ 
    \bottomrule
    \end{tabular}  
    }
    \caption{\textbf{Statistics of our curated CelebV-Dub dataset.}} 
  \label{tab:dataset_statistics}
\end{table}

\paragraph{Speaker classification}
Finally, we classify utterances by speaker identity. Initially, videos extracted from the same original source video are grouped together. However, since we cannot guarantee that all clips from a single video contain the same speaker, we apply an off-the-shelf speaker recognition model~\cite{wavlm} to measure pairwise speaker similarity. Clips within the same original video are re-clustered according to these similarity scores, with a defined threshold determining speaker identity clusters.

\begin{figure}[tp]
    \centering
    \small
    \includegraphics[width=1.0\linewidth]{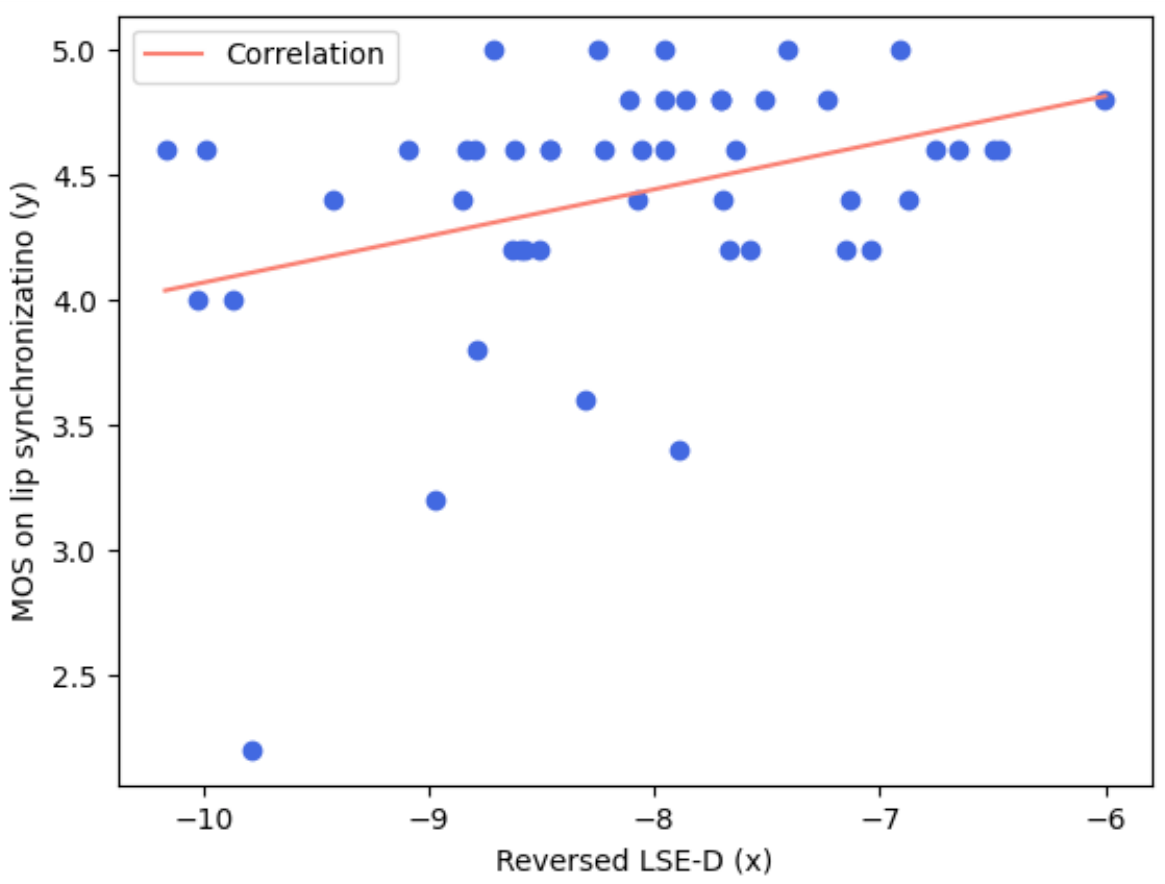}
    \caption{\textbf{Correlation between human evaluation and lip-sync objective metric.} We visualize the scatter plot showing the relationship between the objective lip-sync metric (LSE-D) and the subjective Mean Opinion Score (MOS) on lip-sync from human evaluation. We observe a weak correlation between the two, with a correlation coefficient of 0.36, indicating that LSE-D should be used as a reference rather than a definitive metric.} 
    \label{fig:sync}
\end{figure}

\begin{table*}[tp]
    \centering
    \resizebox{1\linewidth}{!}{\begin{tabular}{lccccccccc}
    \toprule
    \multirow{2}{*}[-0.4em]{A vs. B} & \multicolumn{3}{c}{Naturalness} & \multicolumn{3}{c}{Expressiveness}& \multicolumn{3}{c}{Lip synchronization}\\
    \cmidrule(r{2mm}l{2mm}){2-4} \cmidrule(r{2mm}l{2mm}){5-7} \cmidrule(r{2mm}l{2mm}){8-10}
    & A wins (\%) & Neutral & B wins (\%) & A wins (\%) & Neutral & B wins (\%) & A wins (\%) & Neutral & B wins (\%) \\ 
    \cmidrule{1-10}
    Ours vs. HPMDubbing~\cite{hpmdubbing} & \textbf{99.2} & 0.4 & 0.4 &\textbf{96.4} & 0.4 & 3.2 & \textbf{88.8} & 3.2 & 8.0\\
    Ours vs. StyleDubber~\cite{styledubber} & \textbf{98.0} & 0.4 & 1.6 &\textbf{99.2} & 0.4 & 0.4 & \textbf{91.6} & 6.0 & 2.4\\
    \cmidrule{1-10}
     Ground-Truth vs. HPMDubbing~\cite{hpmdubbing} & 99.6 & 0.0 & 0.4 & 98.8& 0.4 & 0.8 & 89.8& 7.1 & 3.1\\
    Ground-Truth vs. Ours &  \cellcolor{gray!20}58.4 & 24.4 & 17.2 & \cellcolor{gray!20}57.2 & 26.4 & 16.4 & \cellcolor{gray!20}44.0 & 40.0 & 16.0\\
    \bottomrule
    \end{tabular}}
    \caption{\textbf{A/B testing results on CelebV-Dub.} We report the preferences (\%) between A and B across various aspects of synthesized speech. In rows 1 and 2, our model is significantly preferred by humans over existing methods. Comparing rows 3 and 4, HPMDubbing is significantly less preferred compared to the ground truth, while our model, highlighted with a \colorbox{gray!20}{gray background}, is preferred more. Surprisingly, over 41.6\% of the time, our model is perceived as equally good as or better than the ground truth.}
    \label{tab:ab2}
\end{table*}

\begin{table*}[t]
    \centering
    \resizebox{1\linewidth}{!}{\begin{tabular}{lcccccccccc}
    \toprule
    Method &  WER ($\downarrow$)& LSE-D ($\downarrow$)& LSE-C ($\uparrow$) & spkSIM ($\uparrow$) & UTMOS ($\uparrow$) & DNSMOS ($\uparrow$) & MCD ($\downarrow$)& F0 ($\downarrow$) & Energy ($\downarrow$) & emoSIM ($\uparrow$)\\
    \cmidrule{1-11}
    Ground-Truth &5.96 & 7.28 & 7.35 & - & 3.10 & 3.44&  & - & - & -\\
    Zero-shot TTS~\cite{peng2024voicecraft} & 2.97 & 12.52 & 2.08 & 0.279 & 3.02 & 3.54 & - & - & - & 0.733\\
    \cmidrule{1-11}
    HPMDubbing~\cite{hpmdubbing} &17.36 & \textbf{6.98} & \textbf{7.65} & 0.176 & \uline{2.62} & 3.10& 9.11 & 129.30 & 4.87 & 0.759\\
    StyleDubber~\cite{styledubber} &16.06 & 11.38& 3.18 & 0.248 & 2.38 & 3.00& 8.48 & 136.16 & 3.75 & \uline{0.790}\\ 
    Ours (lip-only) &\uline{8.91} & \uline{7.62} & \uline{6.97} & \textbf{0.321} & \textbf{3.55} & \uline{3.51}& \textbf{7.17} & \uline{88.12} & \textbf{2.91} & 0.769\\ 
    Ours (lip $\&$ face) &\textbf{8.11} & 7.71 & 6.88 & \uline{0.312} & \textbf{3.55} & \textbf{3.52}& \uline{7.34} & \textbf{86.01} & \uline{2.94} & \textbf{0.791}\\ 
    \bottomrule
    \end{tabular}}
    \caption{\textbf{Generalization results on Voxceleb2.} We highlight the best results in \textbf{Bold} and \uline{underline} the second best among all the methods.}
    \label{tab:vox}
\end{table*}

\section{Analysis on lip-synchronzation metric}\label{supple:lipsync}
As lip synchronization accuracy (LSE-D) measured by SyncNet~\cite{syncnet} has been shown to be unstable in several studies~\cite{yaman2024audio, ma2025sayanything, yaman2024cvpr}, we investigate whether these findings align within our dataset and cases. Specifically, we analyze the correlation between LSE-D and human evaluation of lip synchronization on the same samples. For this analysis, we use SyncNet to measure the LSE-D for a synthesized speech sample and its corresponding video, while five human evaluators assess the lip synchronization of the same sample using the Mean Opinion Score (MOS). LSE-D is a distance metric, where lower values indicate better lip synchronization, while MOS uses a 1-5 rating scale, with higher values indicating better quality. To facilitate comparison, we reverse the LSE-D score (by taking the negative) to align it with the MOS scale. The MOS for each sample is averaged from the ratings of five human evaluators. A total of 50 samples are used in this analysis.

Figure~\ref{fig:sync} shows a scatter plot with reversed LSE-D on the x-axis and MOS of lip synchronization on the y-axis. Interestingly, we observe a weak correlation between the objective and subjective metrics, with a correlation coefficient of 0.36. Furthermore, even when the LSE-D scores are relatively high (indicating poor lip-sync according to the objective metric), ranging from 7 to 10, human ratings mostly remain above 4, which is considered a relatively high score on the MOS scale. The average LSE-D for the 50 samples in this analysis is 8.11, while the average MOS is 4.42. This suggests that, despite relatively poor LSE-D scores, humans perceive the lip synchronization as sufficiently accurate.

Given the weak correlation between LSE-D and human evaluation, we conclude that human evaluation is the most accurate metric for validating lip synchronization performance. While LSE-D remains a useful objective metric for evaluating lip synchronization, as this analysis shows, it should not be considered definitive; rather, it serves better as a reference metric when human evaluation is limited.

\section{Additional results and analysis}

\begin{figure*}[tp]
    \centering
    \small
    \includegraphics[width=1.0\linewidth]{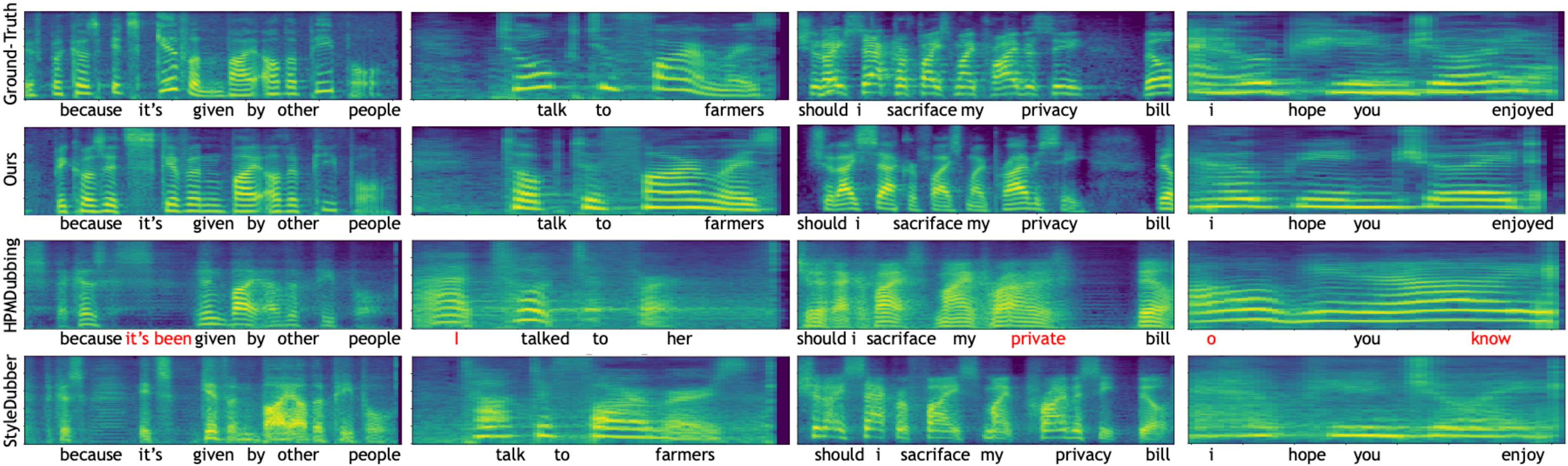}
    \caption{\textbf{Qualitative results.} We compare mel-spectrogram visualizations from ground-truth recordings, our model, and existing methods~\cite{hpmdubbing, styledubber} on the LRS3 (columns 1–2) and our CelebV-Dub (columns 3–4) datasets. The texts below each mel-spectrogram represent time-aligned speech extracted using Whisper~\cite{whisper}, with red text indicating incorrectly synthesized speech.} 
    \label{fig:qual_supple}
\end{figure*}

\subsection{Human evaluation}\label{supple:human}
We present the A/B testing results on our curated CelebV-Dub dataset in \Tref{tab:ab2}. Similar to the main paper, our model is significantly preferred by humans over existing methods, with over 98\% preference for naturalness, 96\% for expressiveness, and 88\% for lip synchronization. Comparing rows 3 and 4, we observe that, in most cases, the ground truth is preferred over HPMDubbing, while our model is rated as good as or better than the ground truth over 41.6\% of the time across all metrics.
These results indicate that, on the CelebV-Dub dataset, which contains expressive content, our model synthesizes speech that is both temporally and semantically aligned with the target video while being sufficiently expressive, leading to high human preference.

It is worth noting that our model performs favorably against the ground truth and outperforms existing methods in lip synchronization during A/B testing, even though our lip synchronization metrics in Sec.~\textcolor{iccvblue}{4.3} of the main paper show lower performance than HPMDubbing~\cite{hpmdubbing}. These results in Sec.~\textcolor{iccvblue}{4.3} may be due to the instability of SyncNet, a limitation discussed in \Sref{supple:lipsync} of the supplementary material and related works~\cite{yaman2024audio, ma2025sayanything, yaman2024cvpr}. Therefore, human evaluation should be given more weight to validate the lip-sync accuracy of the synthesized speech.

\subsection{Generalization results}
We incorporate a subset of the VoxCeleb2~\cite{voxceleb2} dataset to evaluate the generalization performance of our approach. Both our proposed models and the comparison models are trained on the LRS3 dataset~\cite{afouras2018lrs3} and tested on the VoxCeleb2 subset. For this experiment, we select 200 samples from the VoxCeleb2 test split and use the Whisper~\cite{whisper} large model to extract pseudo ground-truth text for each sample.
As summarized in \Tref{tab:vox}, our methods outperform the other approaches across most of the metrics, particularly demonstrating a substantial improvement in WER and automatic MOS evaluations. Both our models perform lower than HPMDubbing on LSE-D/C. However, this can still be considered satisfactory, as the LSE-D scores are better than those in \Sref{supple:lipsync} (average LSE-D of 8.11), which achieved an average MOS of 4.42, indicating sufficiently good lip synchronization.
Interestingly, our model variant using both lip and face input shows a significant improvement in emotional similarity (emoSIM) compared to the lip-only variant, highlighting the advantage of combining both inputs for expressive speech synthesis.

\subsection{Qualitative results}
We visually compare the mel-spectrogram samples synthesized by prior methods~\cite{hpmdubbing, styledubber} and our proposed approach, along with those from the ground-truth recordings in \Fref{fig:qual_supple}. As shown in the results, HPMDubbing often produces incorrect speech, failing to convey accurate content. While StyleDubber performs better than HPMDubbing in terms of content accuracy, its synthesized signal is often blurry, indicating considerable noise, and it frequently exhibits time misalignment (see columns 2 and 3). In contrast, the samples generated by our model accurately convey content with clear and distinct mel frequencies, closely matching the ground-truth mel-spectrograms. These results demonstrate the superiority of our model over existing methods in producing accurate, time-aligned, and high-quality speech.

\section{Details on the objective metrics}
\paragraph{WER}
Word Error Rate (WER) is a widely used metric in the speech-to-text domain. Since the output of automated video dubbing is speech, we rely on an off-the-shelf ASR model to extract text from the synthesized speech and measure the WER. Specifically, we use the Whisper~\cite{whisper} medium.en model to extract text from the synthesized speech and measure WER. 

\paragraph{LSE-D and LSE-C}
To measure lip synchronization accuracy, we assess the audio-visual synchronization between lip movements and speech. We use SyncNet~\cite{syncnet}, which has learned representations for aligning lip movements with corresponding speech snippets. Two metrics are measured using SyncNet: Lip Sync Error - Distance (LSE-D) and Lip Sync Error - Confidence (LSE-C). LSE-D measures the Euclidean distance between the audio and visual embeddings extracted by SyncNet, where lower values indicate better audio-visual synchronization. LSE-C is a probability-based confidence metric derived from the embeddings' distances, with higher values indicating higher confidence in synchronization.

\begin{figure*}[tp]
    \centering
    \includegraphics[width=1\textwidth]{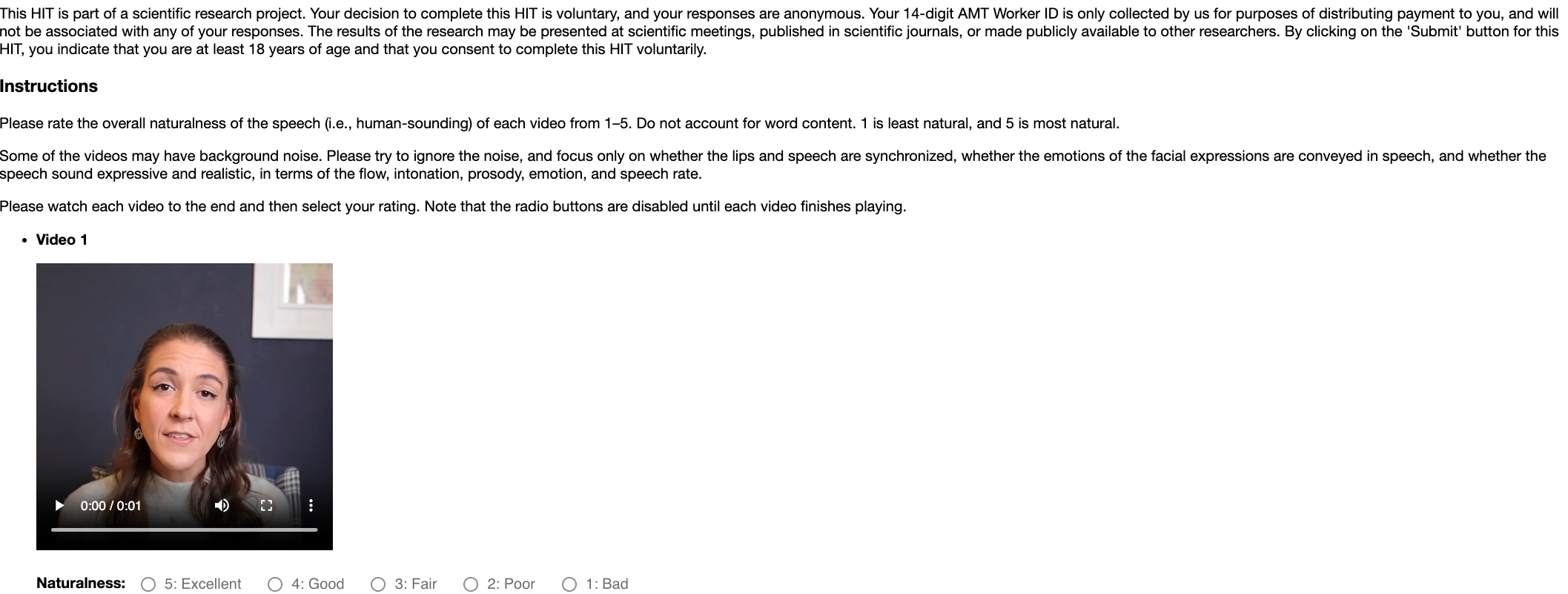}
    \caption{\textbf{Instruction and sample for AMT human listening test on overall naturalness of speech.}}
    \label{fig:amt_nat}
\end{figure*}
\begin{figure*}[tp]
    \centering
    \includegraphics[width=1\textwidth]{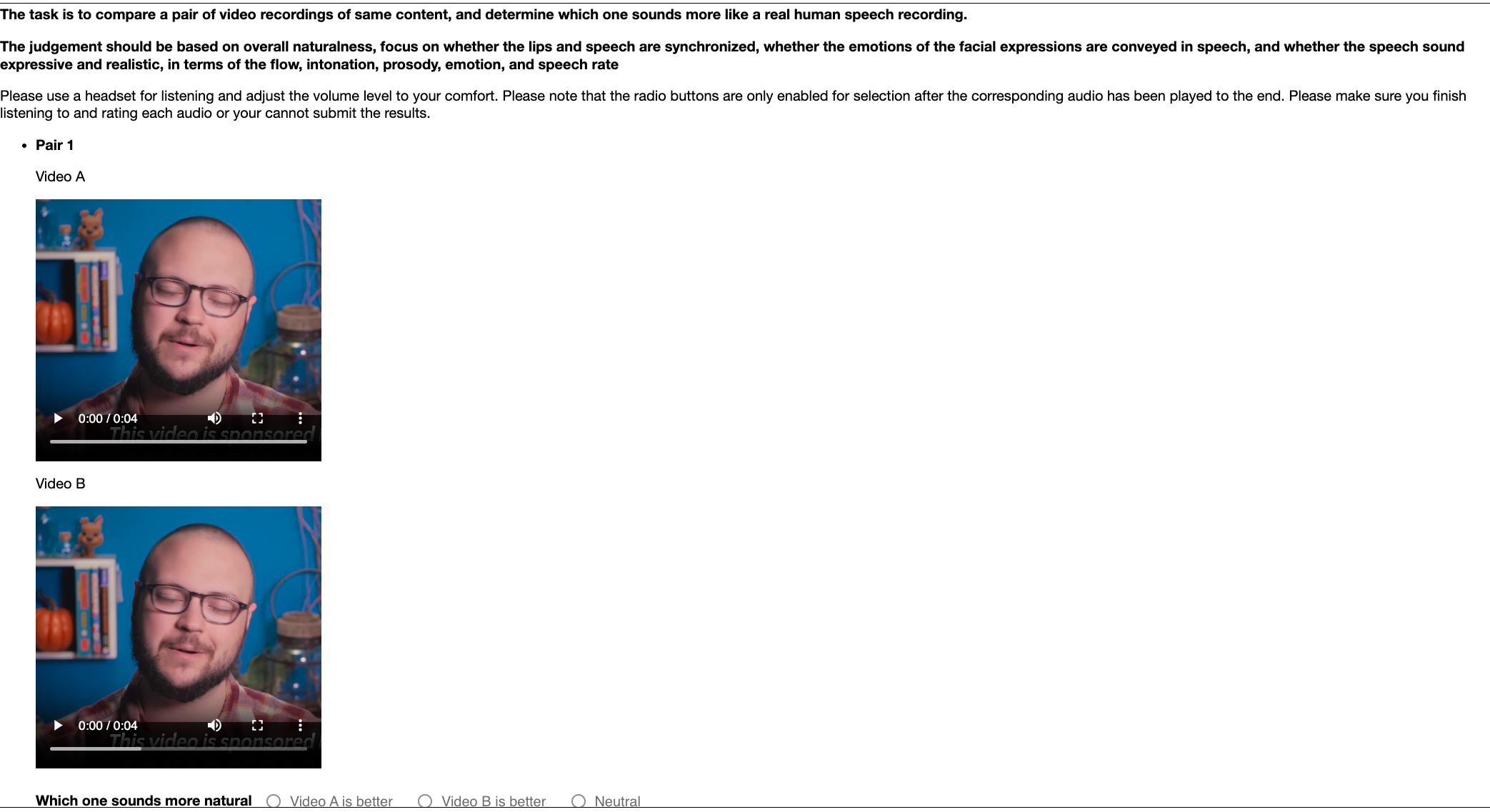}
    \caption{\textbf{Instruction and sample for AMT Human listening test for A/B testing on overall naturalness of speech.}}
    \label{fig:amt_abtest}
\end{figure*}

\paragraph{Speaker similarity (spkSIM)}
We use WavLM-TDNN~\cite{wavlm} to measure speaker similarity. 
As we prompt the source speech from the same speaker as the target speech but a different utterance, we assume the model synthesizes the speech in the target speaker's voice. After synthesizing the speech, we measure the cosine similarity between the synthesized speech features and the ground-truth target speech using the WavLM-TDNN embedding space.

\paragraph{Emotion similarity (emoSIM)}
We measure the expressiveness of the synthesized speech by evaluating the emotion similarity between the synthesized speech and the ground truth. We use the Emotion2Vec~\cite{emofan} model to compute the cosine similarity between the synthesized speech and the ground truth in its embedding space.

\paragraph{DNSMOS and UTMOS}
Deep Noise Suppression MOS (DNSMOS)~\cite{dnsmos} and Universal TTS MOS (UTMOS)~\cite{utmos} are used to objectively evaluate speech quality by approximating subjective human ratings (Mean Opinion Score, MOS). DNSMOS is designed to assess the quality of speech processed by noise suppression algorithms, measuring clarity, naturalness, background noise quality, and overall quality. Similarly, UTMOS focuses on evaluating the quality of synthesized speech, particularly by assessing naturalness, intelligibility, prosody, and expressiveness.

\paragraph{MCD, energy, and F0}
We measure several low-level metrics: Mel-Cepstral Distortion (MCD), F0 distance (F0), and energy distance (Energy). MCD is used to measure the intelligibility of speech, while F0 and Energy are more closely correlated with prosody similarity between the synthesized speech and the ground truth. We follow the implementation of these metrics in \cite{peng2024voicecraft}. MCD measures the difference in Mel Frequency Cepstrum Coefficients (MFCC) between the generated and ground-truth speech, using a 13-order MFCC and the pymcd package for measurement. For F0 measurement, we use the pYIN algorithm~\cite{mauch2014pyin}, implemented in librosa~\cite{mcfee2015librosa}, with minimal and maximal frequencies set to 80Hz and 600Hz, respectively. The energy distance is computed using the root mean square of the magnitude of the spectrogram, extracted via the short-time Fourier transform with a window length of 640 and a hop size of 160.

\section{Additional implementation details}
\paragraph{Training setup}
We introduce two variants of our model in the main paper: one with lip-only input and the other with both lip and face input. We observe that the lip-only variant yields favorable results compared to existing work and ground truth. For training the latter model, we find that starting with the lip-only model and zero-initializing the AV fusion layer for full face input leads to stable training. Furthermore, when training on the CelebV-Dub dataset, we initialize the model with a version pretrained on the LRS3 dataset. We apply the same training setup to existing methods'~\cite{hpmdubbing,styledubber} training to ensure a fair comparison.

\paragraph{Inference setup}
Although our model synthesizes high quality, natural, and lip-synced speech, autoregressive generation may sometimes result in inaccurate output. Therefore, as mentioned in Sec.\textcolor{iccvblue}{4.1} of the paper, we design a sorting strategy similar to VALL-E 2~\cite{valle2}. Given ten synthesized speech samples, ${\mathbf{Y}_{\text{tgt, i}}}^{10}_{i=1}$, we sort them using content accuracy (WER) and lip synchronization accuracy (LSE-D) to select the optimal sample. We denote the WER and LSE-D values for each sample as $\mathbf{Y}_{\text{tgt, i}}^{\text{WER}}$ and $\mathbf{Y}_{\text{tgt, i}}^{\text{LSE-D}}$, respectively. Specifically, we first sort the samples according to LSE-D if the WER is below 5\%, and otherwise, we sort them based on WER, where lower values are preferred. This sorting method is defined as: 
\begin{equation} \mathbf{Y}_{\text{tgt, best}} = \argmin_{\mathbf{Y}_{\text{tgt, i}}}([\min(\mathbf{Y}_{\text{tgt, i}}^{\text{WER}}, 0.05), \mathbf{Y}_{\text{tgt, i}}^{\text{LSE-D}}]).
\end{equation} 
This sorting strategy is also applied to existing methods across all evaluations to ensure a fair comparison.

\section{Details on the human evaluation}
Amazon Mechanical Turk (AMT) is used to conduct human listening tests. We select 100 audio samples from the LRS3 test set and 50 audio samples from the CelebV-Dub test set, totaling 400 samples for LRS3 and 200 samples for CelebV-Dub, with samples from the three models and the ground truth.
For the mean opinion score (MOS), we design an extensive evaluation based on various criteria: naturalness, intelligibility, expressiveness, lip synchronization, and speaker similarity. We use a 5-point Likert scale, where 1 represents ``poor'' and 5 represents ``excellent.''
In the A/B testing, we present two samples to a Turker and ask them to judge which one sounds better in terms of naturalness, expressiveness, or lip synchronization, allowing them to choose either sample as better or neutral. For each sample or comparison, 5 ratings are obtained from different Turkers. We also compute the 95\% confidence interval for MOS.
In the MOS test, 43 Turkers participated in the LRS3 listening test, and 25 Turkers participated in the CelebV-Dub listening test. For A/B testing, 34 Turkers participated in the LRS3 test, and 23 Turkers participated in the CelebV-Dub test.
Please refer to \Fref{fig:amt_nat} and \Fref{fig:amt_abtest} for sample instructions.

%% file: main.bbl
\begin{thebibliography}{54}
\providecommand{\natexlab}[1]{#1}
\providecommand{\url}[1]{\texttt{#1}}
\expandafter\ifx\csname urlstyle\endcsname\relax
  \providecommand{\doi}[1]{doi: #1}\else
  \providecommand{\doi}{doi: \begingroup \urlstyle{rm}\Url}\fi

\bibitem[Afouras et~al.(2018)Afouras, Chung, and Zisserman]{afouras2018lrs3}
Triantafyllos Afouras, Joon~Son Chung, and Andrew Zisserman.
\newblock Lrs3-ted: a large-scale dataset for visual speech recognition.
\newblock \emph{arXiv preprint arXiv:1809.00496}, 2018.

\bibitem[Agostinelli et~al.(2023)Agostinelli, Denk, Borsos, Engel, Verzetti, Caillon, Huang, Jansen, Roberts, Tagliasacchi, et~al.]{agostinelli2023musiclm}
Andrea Agostinelli, Timo~I Denk, Zal{\'a}n Borsos, Jesse Engel, Mauro Verzetti, Antoine Caillon, Qingqing Huang, Aren Jansen, Adam Roberts, Marco Tagliasacchi, et~al.
\newblock Musiclm: Generating music from text.
\newblock \emph{arXiv preprint arXiv:2301.11325}, 2023.

\bibitem[Baade et~al.(2024)Baade, Peng, and Harwath]{baade2024neural}
Alan Baade, Puyuan Peng, and David Harwath.
\newblock Neural codec language models for disentangled and textless voice conversion.
\newblock In \emph{Conference of the International Speech Communication Association (INTERSPEECH)}, 2024.

\bibitem[Bain et~al.(2023)Bain, Huh, Han, and Zisserman]{bain2023whisperx}
Max Bain, Jaesung Huh, Tengda Han, and Andrew Zisserman.
\newblock Whisperx: Time-accurate speech transcription of long-form audio.
\newblock In \emph{Conference of the International Speech Communication Association (INTERSPEECH)}, 2023.

\bibitem[Bernard and Titeux(2021)]{bernard2021phonemizer}
Mathieu Bernard and Hadrien Titeux.
\newblock Phonemizer: Text to phones transcription for multiple languages in python.
\newblock \emph{Journal of Open Source Software}, 2021.

\bibitem[Borsos et~al.(2023)Borsos, Marinier, Vincent, Kharitonov, Pietquin, Sharifi, Roblek, Teboul, Grangier, Tagliasacchi, et~al.]{borsos2023audiolm}
Zal{\'a}n Borsos, Rapha{\"e}l Marinier, Damien Vincent, Eugene Kharitonov, Olivier Pietquin, Matt Sharifi, Dominik Roblek, Olivier Teboul, David Grangier, Marco Tagliasacchi, et~al.
\newblock Audiolm: a language modeling approach to audio generation.
\newblock \emph{IEEE/ACM transactions on audio, speech, and language processing}, 2023.

\bibitem[Chen et~al.(2022)Chen, Wang, Chen, Wu, Liu, Chen, Li, Kanda, Yoshioka, Xiao, et~al.]{wavlm}
Sanyuan Chen, Chengyi Wang, Zhengyang Chen, Yu Wu, Shujie Liu, Zhuo Chen, Jinyu Li, Naoyuki Kanda, Takuya Yoshioka, Xiong Xiao, et~al.
\newblock Wavlm: Large-scale self-supervised pre-training for full stack speech processing.
\newblock \emph{IEEE Journal of Selected Topics in Signal Processing}, 2022.

\bibitem[Chen et~al.(2024)Chen, Liu, Zhou, Liu, Tan, Li, Zhao, Qian, and Wei]{valle2}
Sanyuan Chen, Shujie Liu, Long Zhou, Yanqing Liu, Xu Tan, Jinyu Li, Sheng Zhao, Yao Qian, and Furu Wei.
\newblock Vall-e 2: Neural codec language models are human parity zero-shot text to speech synthesizers.
\newblock \emph{arXiv preprint arXiv:2406.05370}, 2024.

\bibitem[Choi et~al.(2023)Choi, Kim, and Ro]{choi2023intelligible}
Jeongsoo Choi, Minsu Kim, and Yong~Man Ro.
\newblock Intelligible lip-to-speech synthesis with speech units.
\newblock In \emph{Conference of the International Speech Communication Association (INTERSPEECH)}, 2023.

\bibitem[Chung and Zisserman(2017)]{syncnet}
Joon~Son Chung and Andrew Zisserman.
\newblock Out of time: automated lip sync in the wild.
\newblock In \emph{Computer Vision--ACCV 2016 Workshops: ACCV 2016 International Workshops, Taipei, Taiwan, November 20-24, 2016, Revised Selected Papers, Part II 13}, 2017.

\bibitem[Chung et~al.(2018)Chung, Nagrani, and Zisserman]{voxceleb2}
Joon~Son Chung, Arsha Nagrani, and Andrew Zisserman.
\newblock Voxceleb2: Deep speaker recognition.
\newblock In \emph{Conference of the International Speech Communication Association (INTERSPEECH)}, 2018.

\bibitem[Cong et~al.(2023)Cong, Li, Qi, Zha, Wu, Wang, Jiang, Yang, and Huang]{hpmdubbing}
Gaoxiang Cong, Liang Li, Yuankai Qi, Zheng-Jun Zha, Qi Wu, Wenyu Wang, Bin Jiang, Ming-Hsuan Yang, and Qingming Huang.
\newblock Learning to dub movies via hierarchical prosody models.
\newblock In \emph{IEEE Conference on Computer Vision and Pattern Recognition (CVPR)}, 2023.

\bibitem[Cong et~al.(2024)Cong, Qi, Li, Beheshti, Zhang, Hengel, Yang, Yan, and Huang]{styledubber}
Gaoxiang Cong, Yuankai Qi, Liang Li, Amin Beheshti, Zhedong Zhang, Anton Hengel, Ming-Hsuan Yang, Chenggang Yan, and Qingming Huang.
\newblock Styledubber: Towards multi-scale style learning for movie dubbing.
\newblock In \emph{Findings of the Association for Computational Linguistics: ACL 2024}, 2024.

\bibitem[Copet et~al.(2023{\natexlab{a}})Copet, Kreuk, Gat, Remez, Kant, Synnaeve, Adi, and D{\'e}fossez]{copet2023simple}
Jade Copet, Felix Kreuk, Itai Gat, Tal Remez, David Kant, Gabriel Synnaeve, Yossi Adi, and Alexandre D{\'e}fossez.
\newblock Simple and controllable music generation.
\newblock In \emph{Advances in Neural Information Processing Systems (NeurIPS)}, 2023{\natexlab{a}}.

\bibitem[Copet et~al.(2023{\natexlab{b}})Copet, Kreuk, Gat, Remez, Kant, Synnaeve, Adi, and D{\'e}fossez]{copet2024simple}
Jade Copet, Felix Kreuk, Itai Gat, Tal Remez, David Kant, Gabriel Synnaeve, Yossi Adi, and Alexandre D{\'e}fossez.
\newblock Simple and controllable music generation.
\newblock In \emph{Advances in Neural Information Processing Systems (NeurIPS)}, 2023{\natexlab{b}}.

\bibitem[D{\'e}fossez et~al.(2022)D{\'e}fossez, Copet, Synnaeve, and Adi]{defossez2022high}
Alexandre D{\'e}fossez, Jade Copet, Gabriel Synnaeve, and Yossi Adi.
\newblock High fidelity neural audio compression.
\newblock \emph{arXiv preprint arXiv:2210.13438}, 2022.

\bibitem[Garcia et~al.(2023)Garcia, Seetharaman, Kumar, and Pardo]{garcia2023vampnet}
Hugo~Flores Garcia, Prem Seetharaman, Rithesh Kumar, and Bryan Pardo.
\newblock Vampnet: Music generation via masked acoustic token modeling.
\newblock \emph{arXiv preprint arXiv:2307.04686}, 2023.

\bibitem[Hassid et~al.(2022)Hassid, Ramanovich, Shillingford, Wang, Jia, and Remez]{hassid2022more}
Michael Hassid, Michelle~Tadmor Ramanovich, Brendan Shillingford, Miaosen Wang, Ye Jia, and Tal Remez.
\newblock More than words: In-the-wild visually-driven prosody for text-to-speech. 2022 ieee.
\newblock In \emph{IEEE Conference on Computer Vision and Pattern Recognition (CVPR)}, 2022.

\bibitem[Hendrycks and Gimpel(2016)]{gelu}
Dan Hendrycks and Kevin Gimpel.
\newblock Gaussian error linear units (gelus).
\newblock \emph{arXiv preprint arXiv:1606.08415}, 2016.

\bibitem[Hennequin et~al.(2020)Hennequin, Khlif, Voituret, and Moussallam]{spleeter2020}
Romain Hennequin, Anis Khlif, Felix Voituret, and Manuel Moussallam.
\newblock Spleeter: a fast and efficient music source separation tool with pre-trained models.
\newblock \emph{Journal of Open Source Software}, 2020.

\bibitem[Holtzman et~al.(2019)Holtzman, Buys, Du, Forbes, and Choi]{holtzman2019curious}
Ari Holtzman, Jan Buys, Li Du, Maxwell Forbes, and Yejin Choi.
\newblock The curious case of neural text degeneration.
\newblock \emph{arXiv preprint arXiv:1904.09751}, 2019.

\bibitem[Hu et~al.(2021)Hu, Tian, Li, Yuping, Wang, and Zhao]{hu2021neural}
Chenxu Hu, Qiao Tian, Tingle Li, Wang Yuping, Yuxuan Wang, and Hang Zhao.
\newblock Neural dubber: Dubbing for videos according to scripts.
\newblock In \emph{Advances in Neural Information Processing Systems (NeurIPS)}, 2021.

\bibitem[Kain et~al.(2007)Kain, Hosom, Niu, Van~Santen, Fried-Oken, and Staehely]{kain2007improving}
Alexander~B Kain, John-Paul Hosom, Xiaochuan Niu, Jan~PH Van~Santen, Melanie Fried-Oken, and Janice Staehely.
\newblock Improving the intelligibility of dysarthric speech.
\newblock \emph{Speech communication}, 2007.

\bibitem[Kharitonov et~al.(2023)Kharitonov, Vincent, Borsos, Marinier, Girgin, Pietquin, Sharifi, Tagliasacchi, and Zeghidour]{kharitonov2023speak}
Eugene Kharitonov, Damien Vincent, Zal{\'a}n Borsos, Rapha{\"e}l Marinier, Sertan Girgin, Olivier Pietquin, Matt Sharifi, Marco Tagliasacchi, and Neil Zeghidour.
\newblock Speak, read and prompt: High-fidelity text-to-speech with minimal supervision.
\newblock \emph{Transactions of the Association for Computational Linguistics}, 2023.

\bibitem[Kreuk et~al.(2022)Kreuk, Synnaeve, Polyak, Singer, D{\'e}fossez, Copet, Parikh, Taigman, and Adi]{kreuk2022audiogen}
Felix Kreuk, Gabriel Synnaeve, Adam Polyak, Uriel Singer, Alexandre D{\'e}fossez, Jade Copet, Devi Parikh, Yaniv Taigman, and Yossi Adi.
\newblock Audiogen: Textually guided audio generation.
\newblock \emph{arXiv preprint arXiv:2209.15352}, 2022.

\bibitem[Liu et~al.(2023)Liu, Zhang, Lei, Chen, Wang, Li, and Xie]{liu2023promptstyle}
Guanghou Liu, Yongmao Zhang, Yi Lei, Yunlin Chen, Rui Wang, Zhifei Li, and Lei Xie.
\newblock Promptstyle: Controllable style transfer for text-to-speech with natural language descriptions.
\newblock \emph{arXiv preprint arXiv:2305.19522}, 2023.

\bibitem[Lu et~al.(2022)Lu, Sisman, Liu, Zhang, and Li]{lu2022visualtts}
Junchen Lu, Berrak Sisman, Rui Liu, Mingyang Zhang, and Haizhou Li.
\newblock Visualtts: Tts with accurate lip-speech synchronization for automatic voice over.
\newblock In \emph{IEEE International Conference on Acoustics, Speech, and Signal Processing (ICASSP)}, 2022.

\bibitem[Lugaresi et~al.(2019)Lugaresi, Tang, Nash, McClanahan, Uboweja, Hays, Zhang, Chang, Yong, Lee, et~al.]{mediapipe}
Camillo Lugaresi, Jiuqiang Tang, Hadon Nash, Chris McClanahan, Esha Uboweja, Michael Hays, Fan Zhang, Chuo-Ling Chang, Ming~Guang Yong, Juhyun Lee, et~al.
\newblock Mediapipe: A framework for building perception pipelines.
\newblock \emph{arXiv preprint arXiv:1906.08172}, 2019.

\bibitem[Ma et~al.(2025)Ma, Wang, Yang, Hu, Liang, Lin, Li, Meng, et~al.]{ma2025sayanything}
Junxian Ma, Shiwen Wang, Jian Yang, Junyi Hu, Jian Liang, Guosheng Lin, Kai Li, Yu Meng, et~al.
\newblock Sayanything: Audio-driven lip synchronization with conditional video diffusion.
\newblock \emph{arXiv preprint arXiv:2502.11515}, 2025.

\bibitem[Ma et~al.(2023{\natexlab{a}})Ma, Haliassos, Fernandez-Lopez, Chen, Petridis, and Pantic]{vsr}
Pingchuan Ma, Alexandros Haliassos, Adriana Fernandez-Lopez, Honglie Chen, Stavros Petridis, and Maja Pantic.
\newblock Auto-avsr: Audio-visual speech recognition with automatic labels.
\newblock In \emph{IEEE International Conference on Acoustics, Speech, and Signal Processing (ICASSP)}, 2023{\natexlab{a}}.

\bibitem[Ma et~al.(2023{\natexlab{b}})Ma, Zheng, Ye, Li, Gao, Zhang, and Chen]{emotion2vec}
Ziyang Ma, Zhisheng Zheng, Jiaxin Ye, Jinchao Li, Zhifu Gao, Shiliang Zhang, and Xie Chen.
\newblock emotion2vec: Self-supervised pre-training for speech emotion representation.
\newblock \emph{arXiv preprint arXiv:2312.15185}, 2023{\natexlab{b}}.

\bibitem[Mauch and Dixon(2014)]{mauch2014pyin}
Matthias Mauch and Simon Dixon.
\newblock pyin: A fundamental frequency estimator using probabilistic threshold distributions.
\newblock In \emph{IEEE International Conference on Acoustics, Speech, and Signal Processing (ICASSP)}, 2014.

\bibitem[McFee et~al.(2015)McFee, Raffel, Liang, Ellis, McVicar, Battenberg, and Nieto]{mcfee2015librosa}
Brian McFee, Colin Raffel, Dawen Liang, Daniel~PW Ellis, Matt McVicar, Eric Battenberg, and Oriol Nieto.
\newblock librosa: Audio and music signal analysis in python.
\newblock \emph{SciPy}, 2015.

\bibitem[Mira et~al.(2022)Mira, Haliassos, Petridis, Schuller, and Pantic]{mira2022svts}
Rodrigo Mira, Alexandros Haliassos, Stavros Petridis, Bj{\"o}rn~W Schuller, and Maja Pantic.
\newblock Svts: scalable video-to-speech synthesis.
\newblock In \emph{Conference of the International Speech Communication Association (INTERSPEECH)}, 2022.

\bibitem[Nakamura et~al.(2012)Nakamura, Toda, Saruwatari, and Shikano]{nakamura2012speaking}
Keigo Nakamura, Tomoki Toda, Hiroshi Saruwatari, and Kiyohiro Shikano.
\newblock Speaking-aid systems using gmm-based voice conversion for electrolaryngeal speech.
\newblock \emph{Speech communication}, 2012.

\bibitem[Peng et~al.(2024)Peng, Huang, Li, Mohamed, and Harwath]{peng2024voicecraft}
Puyuan Peng, Po-Yao Huang, Shang-Wen Li, Abdelrahman Mohamed, and David Harwath.
\newblock Voicecraft: Zero-shot speech editing and text-to-speech in the wild.
\newblock \emph{arXiv preprint arXiv:2403.16973}, 2024.

\bibitem[Radford et~al.(2023)Radford, Kim, Xu, Brockman, McLeavey, and Sutskever]{whisper}
Alec Radford, Jong~Wook Kim, Tao Xu, Greg Brockman, Christine McLeavey, and Ilya Sutskever.
\newblock Robust speech recognition via large-scale weak supervision.
\newblock In \emph{International Conference on Machine Learning (ICML)}, 2023.

\bibitem[Reddy et~al.(2022)Reddy, Gopal, and Cutler]{dnsmos}
Chandan~KA Reddy, Vishak Gopal, and Ross Cutler.
\newblock Dnsmos p. 835: A non-intrusive perceptual objective speech quality metric to evaluate noise suppressors.
\newblock In \emph{IEEE International Conference on Acoustics, Speech, and Signal Processing (ICASSP)}, 2022.

\bibitem[Ren et~al.(2019)Ren, Ruan, Tan, Qin, Zhao, Zhao, and Liu]{ren2019fastspeech}
Yi Ren, Yangjun Ruan, Xu Tan, Tao Qin, Sheng Zhao, Zhou Zhao, and Tie-Yan Liu.
\newblock Fastspeech: Fast, robust and controllable text to speech.
\newblock In \emph{Advances in Neural Information Processing Systems (NeurIPS)}, 2019.

\bibitem[Saeki et~al.(2022)Saeki, Xin, Nakata, Koriyama, Takamichi, and Saruwatari]{utmos}
Takaaki Saeki, Detai Xin, Wataru Nakata, Tomoki Koriyama, Shinnosuke Takamichi, and Hiroshi Saruwatari.
\newblock Utmos: Utokyo-sarulab system for voicemos challenge 2022.
\newblock \emph{arXiv preprint arXiv:2204.02152}, 2022.

\bibitem[Shi et~al.(2022)Shi, Hsu, Lakhotia, and Mohamed]{avhubert}
Bowen Shi, Wei-Ning Hsu, Kushal Lakhotia, and Abdelrahman Mohamed.
\newblock Learning audio-visual speech representation by masked multimodal cluster prediction.
\newblock In \emph{International Conference on Learning Representations (ICLR)}, 2022.

\bibitem[Sung-Bin et~al.(2024)Sung-Bin, Chae-Yeon, Son, Hyun-Bin, Ju, Nam, and Oh]{sung2024multitalk}
Kim Sung-Bin, Lee Chae-Yeon, Gihun Son, Oh Hyun-Bin, Janghoon Ju, Suekyeong Nam, and Tae-Hyun Oh.
\newblock Multitalk: Enhancing 3d talking head generation across languages with multilingual video dataset.
\newblock In \emph{Conference of the International Speech Communication Association (INTERSPEECH)}, 2024.

\bibitem[Tao et~al.(2021)Tao, Pan, Das, Qian, Shou, and Li]{tao2021someone}
Ruijie Tao, Zexu Pan, Rohan~Kumar Das, Xinyuan Qian, Mike~Zheng Shou, and Haizhou Li.
\newblock Is someone speaking? exploring long-term temporal features for audio-visual active speaker detection.
\newblock In \emph{ACM International Conference on Multimedia (MM)}, 2021.

\bibitem[Toisoul et~al.(2021)Toisoul, Kossaifi, Bulat, Tzimiropoulos, and Pantic]{emofan}
Antoine Toisoul, Jean Kossaifi, Adrian Bulat, Georgios Tzimiropoulos, and Maja Pantic.
\newblock Estimation of continuous valence and arousal levels from faces in naturalistic conditions.
\newblock \emph{Nature Machine Intelligence}, 2021.

\bibitem[Vaswani et~al.(2017)Vaswani, Shazeer, Parmar, Uszkoreit, Jones, Gomez, Kaiser, and Polosukhin]{vaswani2017attention}
Ashish Vaswani, Noam Shazeer, Niki Parmar, Jakob Uszkoreit, Llion Jones, Aidan~N Gomez, {\L}ukasz Kaiser, and Illia Polosukhin.
\newblock Attention is all you need.
\newblock In \emph{Advances in Neural Information Processing Systems (NeurIPS)}, 2017.

\bibitem[Wang et~al.(2023)Wang, Chen, Wu, Zhang, Zhou, Liu, Chen, Liu, Wang, Li, et~al.]{valle1}
Chengyi Wang, Sanyuan Chen, Yu Wu, Ziqiang Zhang, Long Zhou, Shujie Liu, Zhuo Chen, Yanqing Liu, Huaming Wang, Jinyu Li, et~al.
\newblock Neural codec language models are zero-shot text to speech synthesizers.
\newblock \emph{arXiv preprint arXiv:2301.02111}, 2023.

\bibitem[Wang et~al.(2020)Wang, Wu, Song, Yang, Wu, Qian, He, Qiao, and Loy]{wang2020mead}
Kaisiyuan Wang, Qianyi Wu, Linsen Song, Zhuoqian Yang, Wayne Wu, Chen Qian, Ran He, Yu Qiao, and Chen~Change Loy.
\newblock Mead: A large-scale audio-visual dataset for emotional talking-face generation.
\newblock In \emph{European Conference on Computer Vision (ECCV)}, 2020.

\bibitem[Yaman et~al.(2024{\natexlab{a}})Yaman, Eyiokur, B{\"a}rmann, Akti, Ekenel, and Waibel]{yaman2024cvpr}
Dogucan Yaman, Fevziye~Irem Eyiokur, Leonard B{\"a}rmann, Seymanur Akti, Haz{\i}m~Kemal Ekenel, and Alexander Waibel.
\newblock Audio-visual speech representation expert for enhanced talking face video generation and evaluation.
\newblock In \emph{IEEE Conference on Computer Vision and Pattern Recognition (CVPR)}, 2024{\natexlab{a}}.

\bibitem[Yaman et~al.(2024{\natexlab{b}})Yaman, Eyiokur, B{\"a}rmann, Ekenel, and Waibel]{yaman2024audio}
Dogucan Yaman, Fevziye~Irem Eyiokur, Leonard B{\"a}rmann, Haz{\i}m~Kemal Ekenel, and Alexander Waibel.
\newblock Audio-driven talking face generation with stabilized synchronization loss.
\newblock In \emph{European Conference on Computer Vision (ECCV)}, 2024{\natexlab{b}}.

\bibitem[Yang et~al.(2024)Yang, Liu, Huang, Weng, and Meng]{yang2024instructtts}
Dongchao Yang, Songxiang Liu, Rongjie Huang, Chao Weng, and Helen Meng.
\newblock Instructtts: Modelling expressive tts in discrete latent space with natural language style prompt.
\newblock \emph{IEEE/ACM Transactions on Audio, Speech, and Language Processing}, 2024.

\bibitem[Yu et~al.(2023)Yu, Zhu, Jiang, Loy, Cai, and Wu]{celebvtext}
Jianhui Yu, Hao Zhu, Liming Jiang, Chen~Change Loy, Weidong Cai, and Wayne Wu.
\newblock Celebv-text: A large-scale facial text-video dataset.
\newblock In \emph{IEEE Conference on Computer Vision and Pattern Recognition (CVPR)}, 2023.

\bibitem[Zeghidour et~al.(2021)Zeghidour, Luebs, Omran, Skoglund, and Tagliasacchi]{zeghidour2021soundstream}
Neil Zeghidour, Alejandro Luebs, Ahmed Omran, Jan Skoglund, and Marco Tagliasacchi.
\newblock Soundstream: An end-to-end neural audio codec.
\newblock \emph{IEEE/ACM Transactions on Audio, Speech, and Language Processing}, 2021.

\bibitem[Zhang et~al.(2024)Zhang, Li, Cong, Yin, Gao, Yan, Hengel, and Qi]{zhang2024speaker}
Zhedong Zhang, Liang Li, Gaoxiang Cong, Haibing Yin, Yuhan Gao, Chenggang Yan, Anton van~den Hengel, and Yuankai Qi.
\newblock From speaker to dubber: movie dubbing with prosody and duration consistency learning.
\newblock In \emph{ACM International Conference on Multimedia (MM)}, 2024.

\bibitem[Zhu et~al.(2022)Zhu, Wu, Zhu, Jiang, Tang, Zhang, Liu, and Loy]{celebvhq}
Hao Zhu, Wayne Wu, Wentao Zhu, Liming Jiang, Siwei Tang, Li Zhang, Ziwei Liu, and Chen~Change Loy.
\newblock Celebv-hq: A large-scale video facial attributes dataset.
\newblock In \emph{European Conference on Computer Vision (ECCV)}, 2022.

\end{thebibliography}
